\documentclass[runningheads]{llncs}

\usepackage{eccv}

\usepackage{eccvabbrv}

\usepackage{graphicx}
\usepackage{booktabs}
\usepackage{xcolor}
\usepackage{amsmath}
\usepackage{amssymb}
\usepackage{array}
\usepackage{multirow}
\usepackage{color}
\usepackage{colortbl}
\usepackage{framed}
\usepackage{bm}
\usepackage{bbm}
\usepackage{xspace}
\usepackage{enumitem}
\usepackage{xparse}
\usepackage{algorithm}
\usepackage{listings}
\usepackage{url}
\usepackage{pifont}
\usepackage{wrapfig}
\usepackage{adjustbox}

\usepackage[accsupp]{axessibility}  

\definecolor{Gray}{gray}{0.9}
\definecolor{LightCyan}{rgb}{0.88,0.95,1}
\definecolor{blond}{rgb}{0.98, 0.94, 0.75}
\definecolor{teagreen}{rgb}{0.82, 0.94, 0.75}

\def \ie {\emph{i.e.}}
\def \eg {\emph{e.g.}}
\def \etal {\emph{et al.}}

\newcommand{\cmark}{\ding{51}}%
\newcommand{\xmark}{\ding{55}}%

\newcommand{\tit}[1]{\smallbreak\noindent\textbf{#1.}}
\newcommand{\tinytit}[1]{\noindent\textbf{#1.}}

\newcommand{\ours}{CoDE\xspace}
\newcommand{\dataset}{D$^3$\xspace}

\newcommand\blfootnote[1]{%
  \begingroup
  \renewcommand\thefootnote{}\footnote{#1}%
  \addtocounter{footnote}{-1}%
  \endgroup
}

\usepackage{hyperref}

\usepackage{orcidlink}

\begin{document}
\sloppy

\title{Contrasting Deepfakes Diffusion via Contrastive Learning and Global-Local Similarities} 

\titlerunning{Contrasting Deepfakes Diffusion via Contrastive Learning}

\author{Lorenzo Baraldi$^*$\inst{2}\orcidlink{0009-0000-4658-8928} \and
Federico Cocchi$^*$\inst{1,2}\orcidlink{0009-0005-1396-9114} \and
Marcella Cornia\inst{1}\orcidlink{0000-0001-9640-9385} \and \\ 
Lorenzo Baraldi\inst{1}\orcidlink{0000-0001-5125-4957} \and
Alessandro Nicolosi\inst{3}\orcidlink{0009-0007-5071-5687} \and
Rita Cucchiara\inst{1}\orcidlink{0000-0002-2239-283X}
}

\authorrunning{L.~Baraldi et al.}

\institute{University of Modena and Reggio Emilia, Italy
\\
\email{name.surname@unimore.it}
\and
University of Pisa, Italy
\\
\email{name.surname@phd.unipi.it}
\and
Leonardo S.p.A.
\\
\email{name.surname@leonardo.com}
}

\maketitle

\begin{abstract}
Discerning between authentic content and that generated by advanced AI methods has become increasingly challenging. While previous research primarily addresses the detection of fake faces, the identification of generated natural images has only recently surfaced. This prompted the recent exploration of solutions that employ foundation vision-and-language models, like CLIP. However, the CLIP embedding space is optimized for global image-to-text alignment and is not inherently designed for deepfake detection, neglecting the potential benefits of tailored training and local image features. In this study, we propose CoDE (Contrastive Deepfake Embeddings), a novel embedding space specifically designed for deepfake detection. CoDE is trained via contrastive learning by additionally enforcing global-local similarities. To sustain the training of our model, we generate a comprehensive dataset that focuses on images generated by diffusion models and encompasses a collection of 9.2 million images produced by using four different generators. Experimental results demonstrate that CoDE achieves state-of-the-art accuracy on the newly collected dataset, while also showing excellent generalization capabilities to unseen image generators.
Our source code, trained models, and collected dataset are publicly available at: \url{https://github.com/aimagelab/CoDE}.
\blfootnote{$^*$Equal contribution.}
\end{abstract}

\setcounter{footnote}{0}
\section{Introduction\vspace{-0.1cm}}
\label{sec:intro}
Thanks to the increasing generation quality of text-to-image models~\cite{rombach2022high}, the dissemination of generated visual content has raised concerns about their use for malicious purposes. Indeed, the threat of a scenario in which visually indistinguishable fake images can be easily generated from simple textual descriptions is evolving. In this context, deepfake detection methods~\cite{ojha2023towards} play a significant role in safeguarding society from the risky scenario in which real and fake images are no longer distinguishable from the naked eye.

Deepfake detection approaches have been extensively studied for the case of AI-manipulated faces~\cite{ijcai2020p476,rossler2019faceforensics++,yang2019exposing}, finding common generation imprints~\cite{wang2020cnn} among different Generative Adversarial Networks (GANs)~\cite{goodfellow2014generative,karras2017progressive}. However, with the advent of diffusion models~\cite{sohl2015deep}, the landscape of image generation has undergone a profound transformation in terms of generation quality and detection techniques. Recent deepfake detection proposals~\cite{ojha2023towards,amoroso2023parents} have considered the adoption of general-purpose vision-and-language backbones like CLIP~\cite{radford2021learning}, and have shown their capability of discriminating images and generalize across generators, including diffusion-based ones.

\begin{figure}[t]
    \centering
    \includegraphics[height=0.215\linewidth]{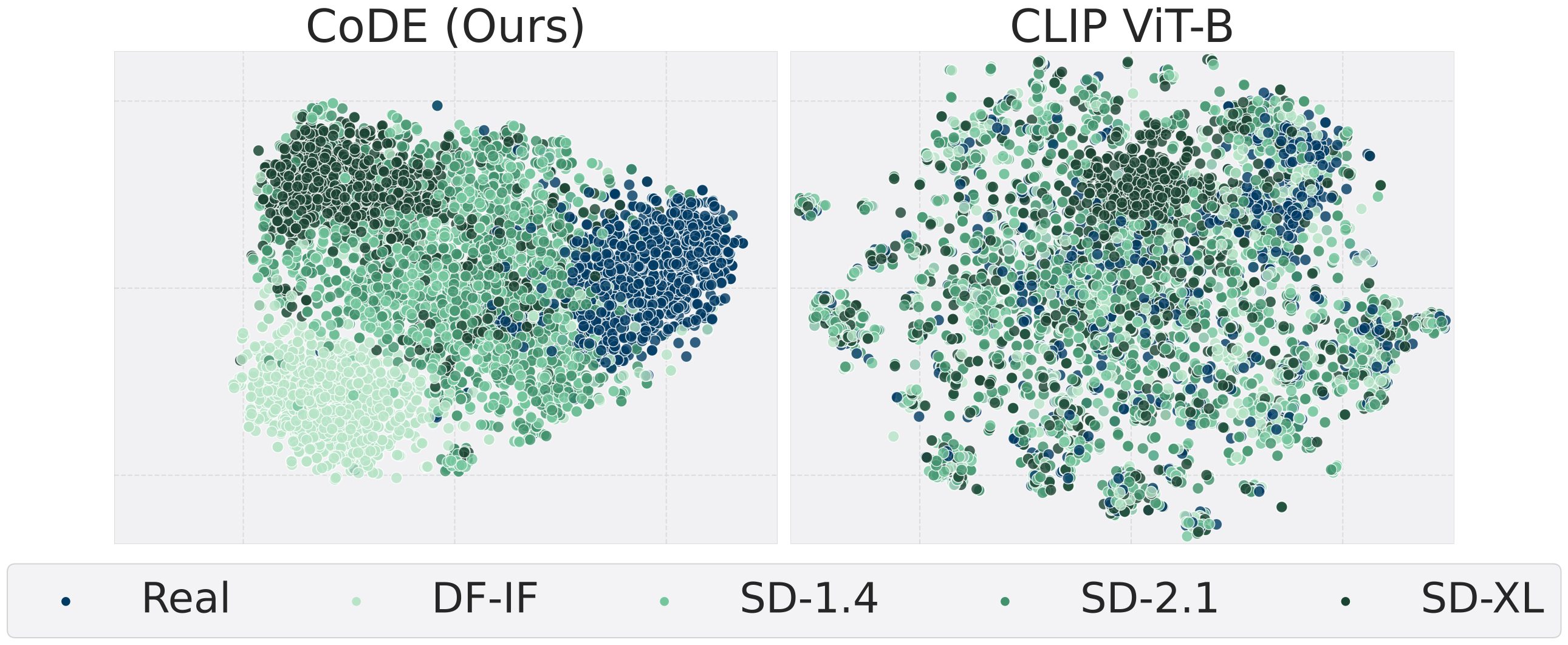}
    \includegraphics[height=0.215\linewidth]{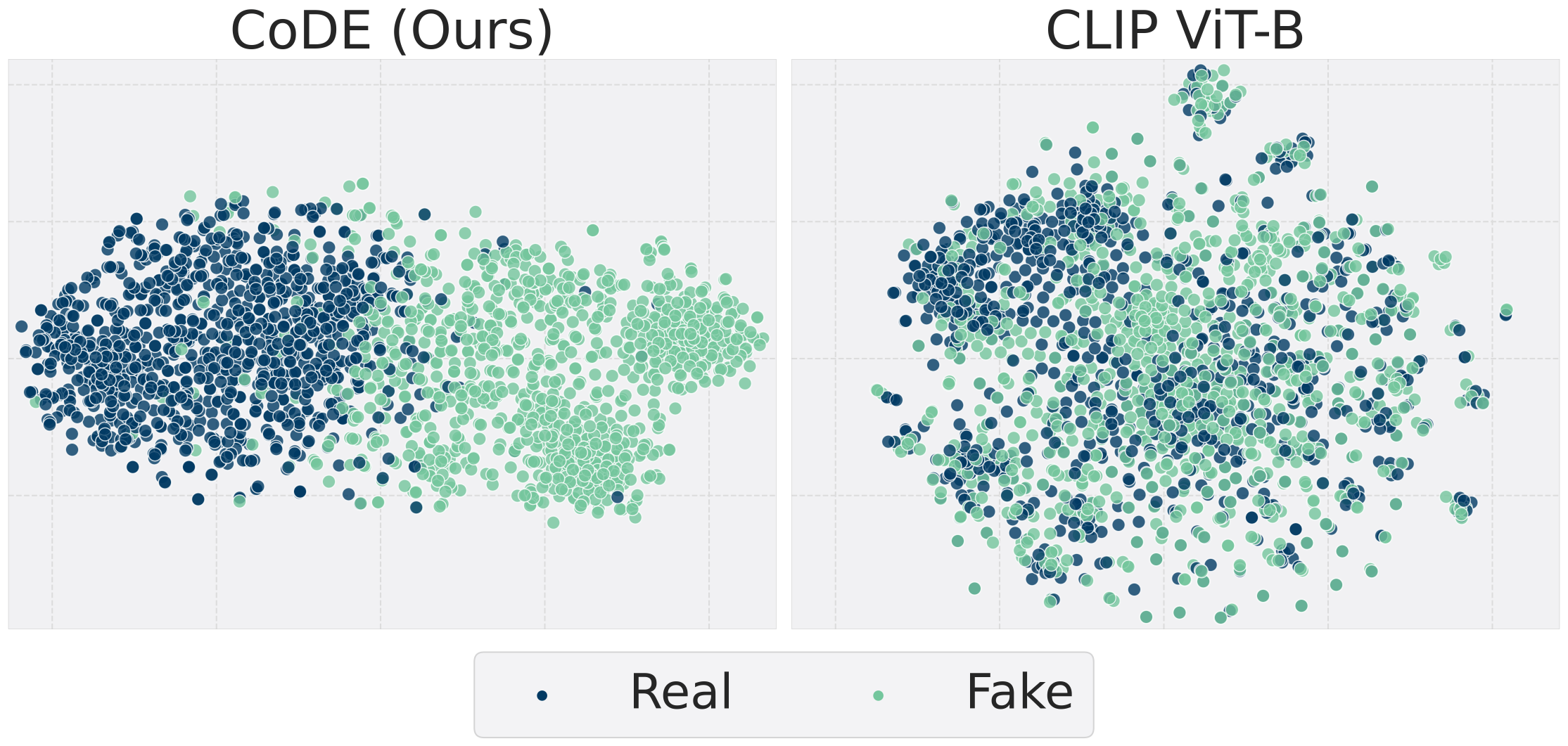}
    \vspace{-.2cm}
    \caption{t-SNE embedding visualization of \ours (left) and CLIP ViT-B (right), considering a real-fake binary representation and a per generator representation. \ours provides tailored and effective features for deepfake classification.}
    \label{fig:first_page}
    \vspace{-.45cm}
\end{figure}

However, the CLIP embedding space has been trained on \textit{real images} only, and enforcing \textit{global} text-to-image similarities through contrastive learning. Clearly, training on real images only and treating the ``fake'' class as an outlier can improve generalization across different generators, but also comes at the cost of limiting the recognition accuracy with respect to a training scenario in which visual features can be learned on fake images as well. Further, generated images can be distinguished prominently by looking at local and fine-grained details, rather than exclusively considering the global context of the image, making the CLIP training paradigm sub-optimal in this context. Lastly, employing pre-trained backbones prevents any architectural deviation from already existing models. As CLIP backbones employ at least 86 million parameters (in the ViT Base~\cite{dosovitskiy2020image} configuration), this also has a significant impact on their adoption in production-ready scenarios where deepfake detection should run in real-time.

In light of these considerations, in this paper we propose \textbf{Co}ntrastive \textbf{D}eepfake \textbf{E}mbeddings (\textbf{\ours}), a novel contrastive-based embedding space specifically tailored for distinguishing real and generated images. Our model is trained by employing both real and fake images, and by stimulating the learning of local-to-global correspondences between real and fake counterparts and between same-class images, in addition to enforcing global similarities. Further, by training from scratch on the real-fake distribution, we also reduce the scale of the model and find that a ViT Tiny architecture can provide state-of-the-art results while being significantly lighter in computational terms. The resulting embedding space can better separate between real and fake samples, and also better distinguish among different generators, as shown in Fig.~\ref{fig:first_page}.

Clearly, training from scratch while maintaining good generalization capabilities requires employing data in both quantity, quality, and diversity. Existing detection datasets are predominantly centered around GAN generators, with only a limited number of datasets~\cite{amoroso2023parents,ojha2023towards} providing a modest volume of images generated by a single diffusion model. Consequently, we generate and release the \textbf{D}iffusion-generated \textbf{D}eepfake \textbf{D}etection (\textbf{D$^3$}) dataset containing 2.3 million records, each composed of a real image coming from LAION-400M~\cite{schuhmann2021laion} dataset and images from four generators, for a total of 9.2 million generated images. To verify the generation capabilities of deepfake detection methods to unseen generators, we also collect a challenging test set composed of 4.8k real images, each paired with 12 fake images generated by as many diffusion-based generators.

We empirically show that \ours can overtake the results obtained by pre-trained models and previous approaches on deepfake detection, while having a smaller amount of learnable parameters. For example, \ours surpasses a CLIP ViT-L model on the D$^3$ test set without external generators by 5.5\% and 4.8\% in accuracy, respectively when using linear and nearest neighbor classifiers.
In a more real-world scenario in which deepfake detectors are tested on generators not seen during training, \ours still achieves the best results compared to different baselines and competitors. In this setting, pairing our backbone with a one-class SVM classifier leads to better and more robust results with an average accuracy over fake images of 94.7\% on the D$^3$ test set extended with fake images from 12 generators. Following previous works~\cite{ojha2023towards}, we extend our analysis to an additional mixture of diffusion and autoregressive generators which are not represented in D$^3$. In this analysis, we also include images generated by recent non-public and commercial tools such as DALL-E 2~\cite{ramesh2022hierarchical}, DALL-E 3~\cite{betker2023improving}, and Midjourney. Even in this setting, \ours confirms its robustness to generators unseen during training achieving state-of-the-art performance compared to existing methods.

To sum up, the main contributions of this work are the development of a new contrastive-based embedding space for deepfake detection. Additionally, we generate and openly release a new dataset comprising over 9 million images produced by diverse diffusion-based generators. We demonstrate through extensive experiments and analysis the effectiveness of the proposed solution, which outperforms existing methods in recognizing deepfakes generated by a wide variety of diffusion-based models.

\section{Related Work\vspace{-0.1cm}}
\label{sec:related}

\tinytit{Image generation models}
Images can be generated with different approaches, ranging from autoregressive models~\cite{esser2021taming,ramesh2021zero,ding2021cogview,yu2022scaling} and generative adversarial networks~\cite{liao2022text,tao2023galip,choi2018stargan,brock2018large,karras2017progressive} to diffusion models~\cite{ho2020denoising,sohl2015deep,balaji2022ediffi}.
The emergence of latent diffusion models~\cite{rombach2022high,dhariwal2021diffusion,nichol2021glide,sauer2023adversarial,podell2023sdxl,razzhigaev2023kandinsky} has significantly propelled the adoption of diffusion-based generators. This is attributed to the heightened efficiency observed in both training and inference, achieved by transitioning the generation process from pixel space to latent space. Notably, this transition maintains state-of-the-art performance in the quality of generated images.
Furthermore, numerous endeavors in literature are directed towards improving both image quality and inference time even further~\cite{patil2024amused,chen2023pixart,hong2023improving}.
In contrast, Imagen~\cite{saharia2022photorealistic} proposes an approach that works directly on the pixel space. Their model initiates the generation process with a lower-resolution image, which is subsequently upsampled through the employment of two text-conditional super-resolution diffusion models. Although this approach yields highly realistic images, it also features higher computational requirements. Given the success and realism of diffusion models, in this paper we will focus on these types of generators.

\tit{Fake image detection}
The detection of generated images has constituted an active area of research in the past few years. Initial works~\cite{rossler2019faceforensics++,yang2019exposing,ijcai2020p476} have focused on the detection of synthetic faces and GAN generators. Subsequently, different studies~\cite{wang2020cnn,cozzolino2018forensictransfer,gragnaniello2021gan} have explored the generalization capabilities of detection models in zero-shot scenarios, where unseen generators are encountered at test time. These researches indicate that diverse GAN generators share common discriminative clues. On the same line, Frank~\etal~\cite{frank2020leveraging} has conducted investigations in the frequency domain, reporting distinctly different spectral features between real and fake images.

With the advent of diffusion models, several works have focused on them. While these generators exhibit distinctive generation imprints~\cite{Corvi_2023_CVPR}, recent research~\cite{corvi2023detection} has shown that detectors trained on GANs generalize poorly on other types of generators. In response to these limitations, recent works~\cite{amoroso2023parents,cocchi2023unveiling,sha2022fake,ojha2023towards,epstein2023online} have designed approaches specifically tailored for diffusion models. A common thread is the utilization of CLIP~\cite{radford2021learning} as the basis feature space. A distinctive approach has been undertaken by Wang~\etal~\cite{wang2023dire}, who has introduced a novel deepfake detection pipeline that works on the difference between the input image and its reconstruction obtained by a pre-trained diffusion model.
Noticeably, these approaches primarily leverage pre-trained foundation models which are not explicitly designed for deepfake detection. In contrast, our proposal focuses on constructing a compact embedding space tailored specifically for this task.

\section{Proposed Method\vspace{-0.1cm}}
\label{sec:method}
\begin{figure*}[t]
    \centering
    \includegraphics[width=\linewidth]{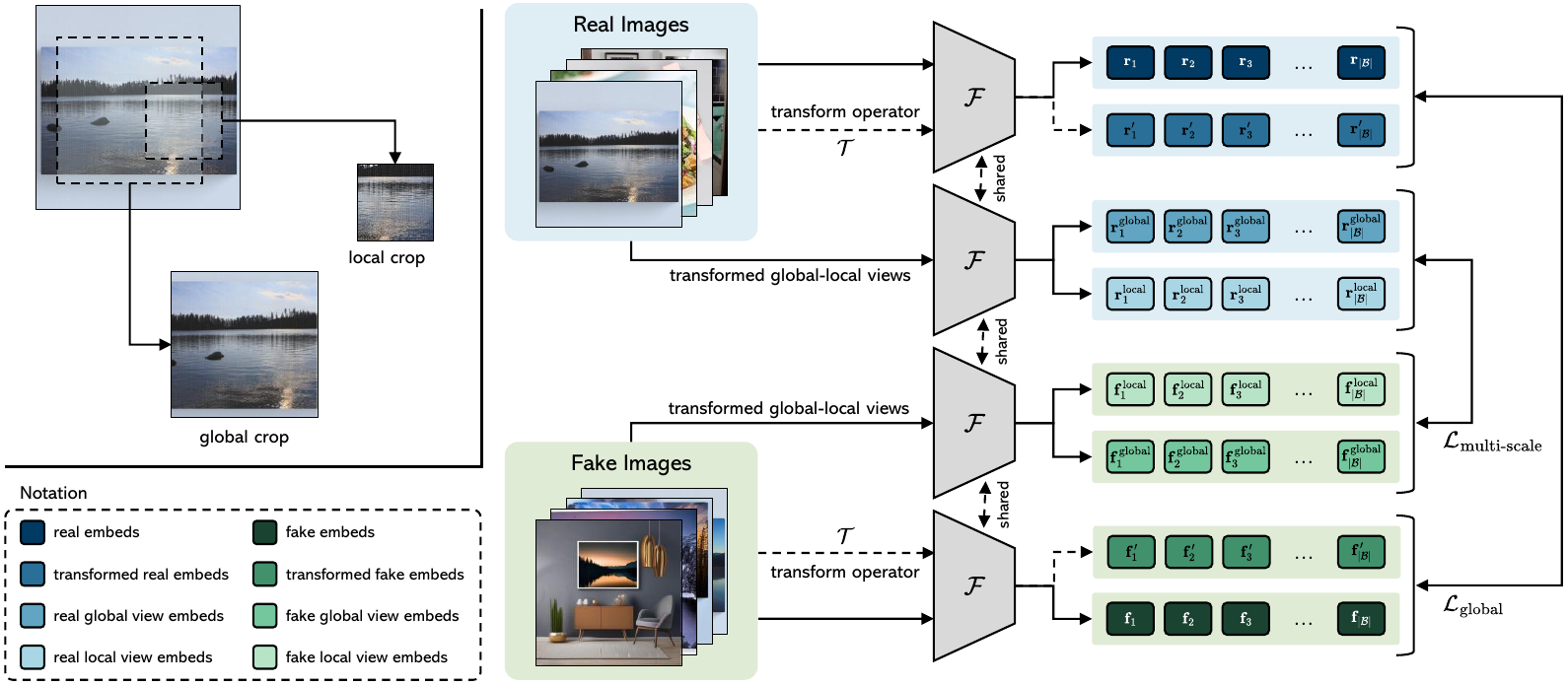}
    \vspace{-.55cm}
    \caption{Visual representation of local and global crops of an input image (left), and overview of \ours (right). Our embedding space is trained by ensuring alignment between local and global crops.}
    \label{fig:model}
    \vspace{-.45cm}
\end{figure*}

Given an input image, we tackle the task of classifying whether it was actually produced by a camera (real image) or whether it was generated by a generative model (fake image). In \ours, we train a contrastive space for deepfake detection with a pair of losses that encourage features coming from real and fake images to be separated in the embedding space, while considering both global and local visual cues. An overview of our approach is shown in Fig.~\ref{fig:model}.

\tit{An embedding space for deepfake detection}
Our contrastive pre-training objective relies on the Info-NCE framework~\cite{oord2018representation}. It utilizes a large collection of fake images generated from web-scraped prompts, paired with a collection of real images having the same size. Given a minibatch $\mathcal{B} = \left( \{R_i\}_{i=1}^N,  \{F_i\}_{i=1}^N \right)$, where $\{R_i\}_{i=1}^N$ and $\{F_i\}_{i=1}^N$ denote, respectively, randomly sampled sets of real and fake images, the contrastive objective encourages real and fake images to lie separated in a shared embedding space, while ensuring invariance to image transformation.
Images $R_i$ and $F_i$ are processed by a learnable Vision Transformer~\cite{dosovitskiy2020image} $\mathcal{F}$ to get their feature embeddings. These are then normalized by their $\ell_2$ norm to get $\mathbf{r}_i = \frac{\mathcal{F}(R_i)}{\|\mathcal{F}(R_i)\|_2} \in \mathbb{R}^d$ and $\mathbf{f}_i = \frac{\mathcal{F}(F_i)}{\|\mathcal{F}(F_i)\|_2} \in \mathbb{R}^d$, where $d$ is the dimensionality of the shared embedding space. The dot product between $\mathbf{r}_i$ and $\mathbf{f}_j$ computes their cosine similarity and accounts for their similarity in the embedding space. 

In contrast to previous literature which tends to leverage pre-trained backbones, we train $\mathcal{F}$ from scratch and exclusively for the task of detecting generated images. This gives us additional flexibility in architectural terms and ultimately allows us to employ a smaller and more efficient backbone, which advantages its application in production-ready scenarios.

\tit{Imposing robustness to transformations}
To ensure robustness to image transformations, we also define a transform operator $\mathcal{T}(\cdot)$ which samples random transformation types from a pre-defined set of operators (\eg~resize, blurring, rotation, etc.) and applies them to an input image at training time. In addition to sampling transformation types, the operator randomly samples the number of transformations in the chain and their strength (\eg~the number of degrees of a rotation, or the area ratio of a crop with respect to the input image). As a result, an image to which $\mathcal{T}$ is applied simulates a transformation process with a variable number of transformations, of variable type and strength.

With the objective of realistically simulating the manipulations that an image can encounter, we include a comprehensive set of transformations: (\textit{i}) blurring; (\textit{ii}) alteration of brightness, contrast, saturation, sharpness and opacity; (\textit{iii}) pixelization, padding, rotation, horizontal flip, aspect ratio change, resize and scale change, skew on the $x$ or $y$ axis; (\textit{iv}) reduction of the JPEG encoding quality level; (\textit{v}) transformation to grayscale; (\textit{vi}) overlay of striped patterns. Additional details on the transformation protocol are reported in the supplementary.

\tit{Separating real and fake samples}
On the basis of the InfoNCE paradigm and of the transformation protocol outlined above, we define a loss for deepfake detection which aims at imposing a separation between real and fake samples in the shared embedding space, starting from embeddings of entire images. Given a batch consisting of real and fake images in equal quantity, we augment each of them by employing the transform operator $\mathcal{T}$. We respectively call $\mathbf{f}'_i$ the feature vector of a transformed fake image $\mathcal{T}(F_i)$ and $\mathbf{r}'_i$ the feature vector of a transformed real image $\mathcal{T}(R_i)$. 

Given a real sample $\mathbf{r}_i$ our loss function aims at maximizing its similarity with a randomly chosen augmented sample $\mathbf{r}'_z$, where $z \in \mathbb{N}_N -\{ i\}$ and minimizing its similarity with respect to all augmented fake samples in the minibatch $\{\mathbf{f}'_j\}_{j=1}^N$. The same objective is applied, symmetrically, when considering fake instances. Each fake sample $\mathbf{f}_i$ is attracted to a randomly chosen augmented sample $\mathbf{f}'_z$, and repulsed from all augmented real samples in the minibatch $\{\mathbf{r}'_j\}_{j=1}^N$. As this objective is applied to embeddings obtained from entire images, the loss operates on a global scale of the image. Formally, our loss function is defined as a pair of cross-entropy losses as follows:
\begin{align}\label{eq:global}
    \mathcal{L}_\text{global} = - \frac{1}{ 2|\mathcal{B} | } \sum_{i=1}^{\mathcal{B}} 
    \left(
        \overbrace{
            \log \frac{e^{t\mathbf{r}_i \cdot \mathbf{r}'_z}}{
                e^{t\mathbf{r}_i \cdot \mathbf{r}'_z} + \sum_{j=1}^{\mathcal{B}} e^{t\mathbf{r}_i \cdot \mathbf{f}'_j}} +}^{\text{real }\rightarrow\text{ fake softmax}}  \right. 
        \left. \underbrace{
            \log \frac{e^{t\mathbf{f}_i \cdot \mathbf{f}'_z}}{
                e^{t\mathbf{f}_i \cdot \mathbf{f}'_z} + \sum_{j=1}^{\mathcal{B}} e^{t\mathbf{f}_i \cdot \mathbf{r}'_j}}}_{\text{fake }\rightarrow\text{ real softmax}}
    \right) \\ \nonumber \text{with } z \neq i,
\end{align}

\noindent where $\cdot$ indicates the dot product and $t$ is a fixed temperature hyper-parameter that controls the peakness of the probability distribution of the loss function.

\tit{Learning multi-scale contrastive features}
The aforementioned contrastive loss requires learning embeddings of the entire image which are appropriate for deepfake detection. However, this objective alone does not explicitly focus on learning good local and multi-scale visual features. Detecting a generated image, indeed, is not exclusively a matter of extracting features from the image as a whole, but is also a setting where local image features play an important role. 

Following this insight, we build a loss component that acts in a multi-scale manner. To this end, we construct different crops of both real and fake images with a multi-crop strategy. Noticeably, multi-crop strategies have been popularized by self-supervised methods like DINO~\cite{caron2020unsupervised,caron2021emerging}, which employ a teacher and a student network and encourage local-to-global correspondences by feeding global views to the teacher and local views to the student. In our case, instead, both global and local views are passed to the same network which, together with a contrastive objective, is in charge of learning proper ``local-to-global'' correspondences shared across samples of the same category (\ie~across different real or fake samples) and discriminative ``local-to-global'' differences between different categories. Computationally, this also has the advantage of training a single backbone rather than training two backbones at once.

In particular, we extract global scale and local scale views of smaller resolutions. Both crops are passed through the embedding network $\mathcal{F}$ to get their embeddings. We respectively call $\mathbf{f}_i^{\text{local}}$ and $\mathbf{f}_i^{\text{global}}$ the embeddings of the local scale and global scale crop of a fake image $F_i$, and $\mathbf{r}_i^{\text{local}}$ and $\mathbf{r}_i^{\text{global}}$ the embeddings of the local and global crop of a real image $R_i$. For improved robustness, also local and global crops are augmented using the transform operator $\mathcal{T}$. The loss function is then defined as
\begin{align}\label{eq:multiscale}
    \mathcal{L}_\text{multi-scale} = - \frac{1}{ 2|\mathcal{B} | } \sum_{i=1}^{\mathcal{B}} 
    \Biggl(&\log \frac{e^{t\mathbf{r}_i^{\text{local}} \cdot \mathbf{r}^{\text{global}}_z}}{
                e^{t\mathbf{r}_i^{\text{local}} \cdot \mathbf{r}^{\text{global}}_z} + \sum_{j=1}^{\mathcal{B}} e^{t\mathbf{r}^{\text{local}}_i \cdot \mathbf{f}^{\text{global}}_j}} +  \\ \nonumber
            &\log \frac{e^{t\mathbf{f}^{\text{local}}_i \cdot \mathbf{f}^{\text{global}}_z}}{
                e^{t\mathbf{f}^{\text{local}}_i \cdot \mathbf{f}^{\text{global}}_z} + \sum_{j=1}^{\mathcal{B}} e^{t\mathbf{f}^{\text{local}}_i \cdot \mathbf{r}^{\text{global}}_j}}
    \Biggl) \quad \text{with } z \neq i.
\end{align}

We follow the standard setting for multi-crop by using global views at resolution $224^2$ covering a large area of the original image, and local views of resolution $96^2$ covering small areas (less than 50\%) of the original image. 

\tit{Test protocol}
Once the embedding space has been trained, we predict the class of an input image (\ie~real or fake) according to three protocols: a nearest neighbor approach, a linear classification strategy, and a one-class SVM approach. 
In the nearest neighbor, we use the trained visual encoder to map the entire training set to its embedding representation, building a bank of real and fake embeddings. At test time, an input image is firstly projected into the same embedding space, and then cosine distance is applied as the metric we find its nearest neighbor in the training set. The prediction (\ie~the output class) is then the same class of the nearest training element found in the bank. 
In linear classification, instead, we add a single linear layer with sigmoid activation to our embedding space, and train only this new classification layer for real-vs-fake classification, with a binary cross-entropy loss. 
When using a one-class SVM classifier, we fit only on real images and treat fake images as outliers residing beyond the boundaries delineated by the classifier with a polynomial kernel.

\section{The D$^3$ Dataset\vspace{-0.1cm}}
\label{sec:dataset}
\begin{figure}[t]
\centering
\large
\setlength{\tabcolsep}{.2em}
\resizebox{\linewidth}{!}{
\begin{tabular}{ccccc c ccccc}
\textbf{Real} & \textbf{DF-IF} & \textbf{SD-1.4} & \textbf{SD-2.1} & \textbf{SD-XL} & & \textbf{Real} & \textbf{DF-IF} & \textbf{SD-1.4} & \textbf{SD-2.1} & \textbf{SD-XL} \\
\addlinespace[0.08cm]
\includegraphics[width=0.195\linewidth]{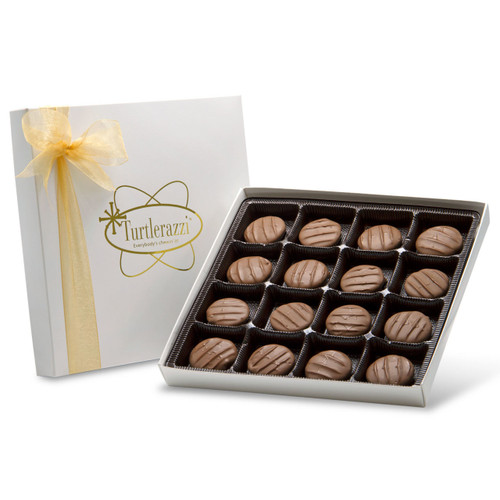} & 
\includegraphics[width=0.195\linewidth]{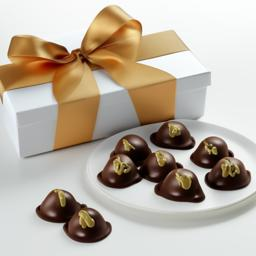} & 
\includegraphics[width=0.195\linewidth]{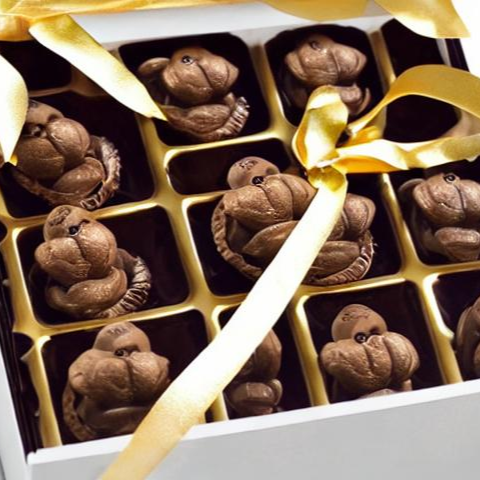} & 
\includegraphics[width=0.195\linewidth]{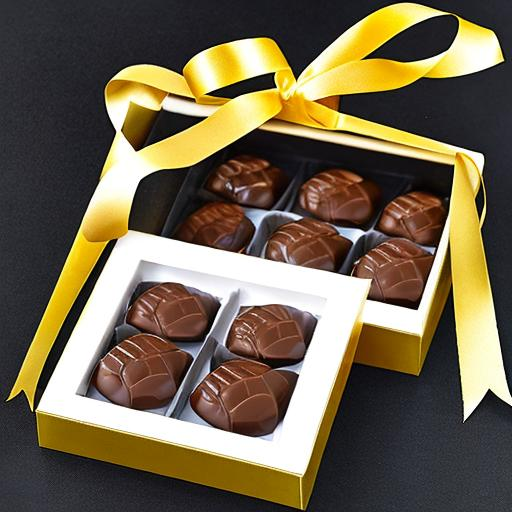} & 
\includegraphics[width=0.195\linewidth]{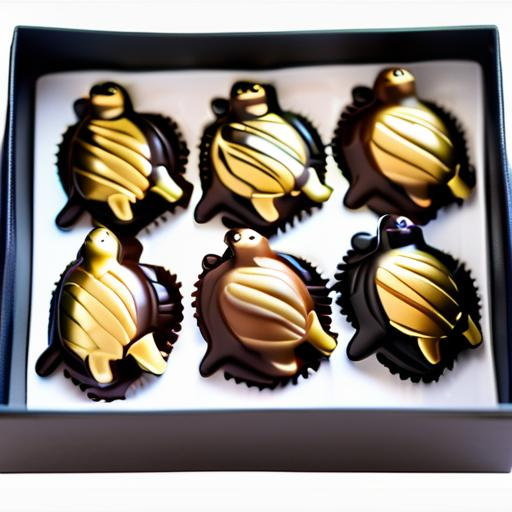} & & 
\includegraphics[width=0.195\linewidth]{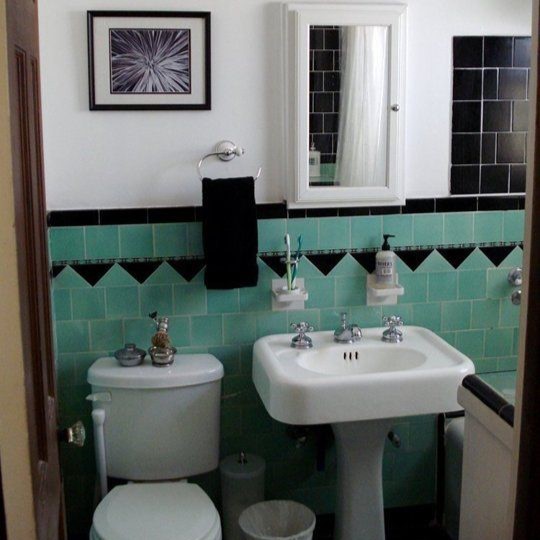} & 
\includegraphics[width=0.195\linewidth]{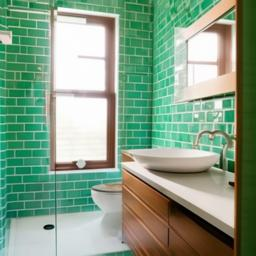} & 
\includegraphics[width=0.195\linewidth]{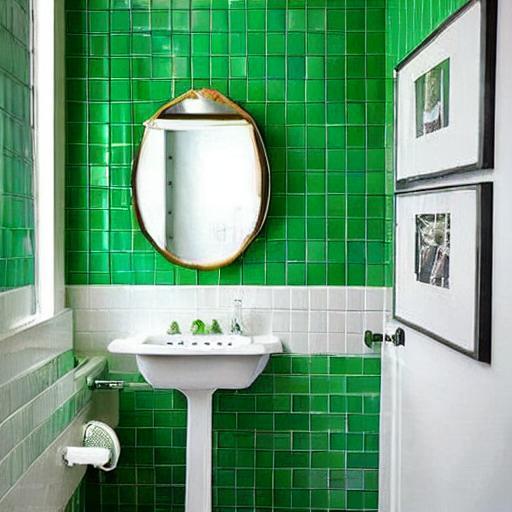} & 
\includegraphics[width=0.195\linewidth]{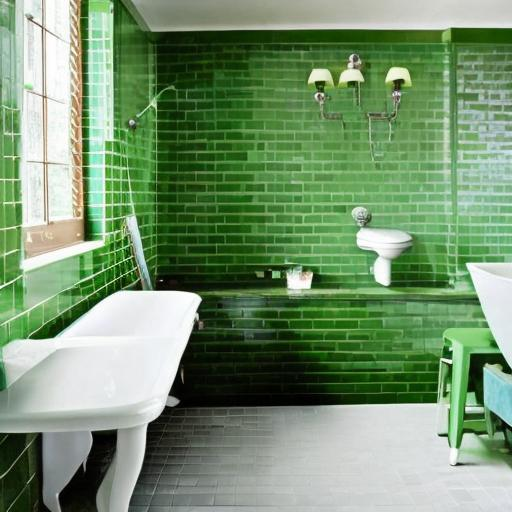} & 
\includegraphics[width=0.195\linewidth]{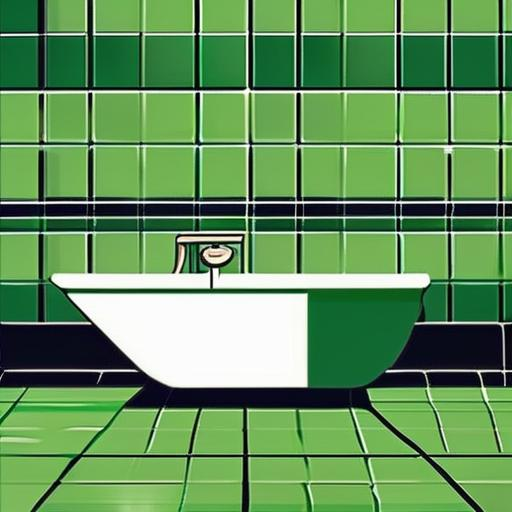} \\
\includegraphics[width=0.195\linewidth]{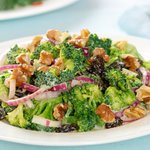} & 
\includegraphics[width=0.195\linewidth]{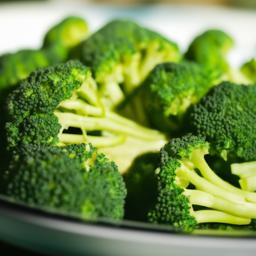} & 
\includegraphics[width=0.195\linewidth]{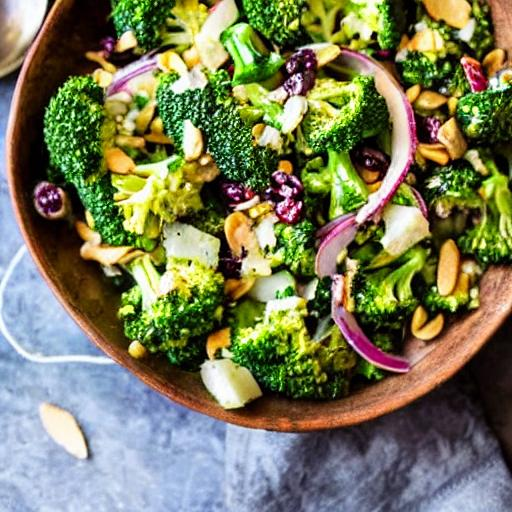} & 
\includegraphics[width=0.195\linewidth]{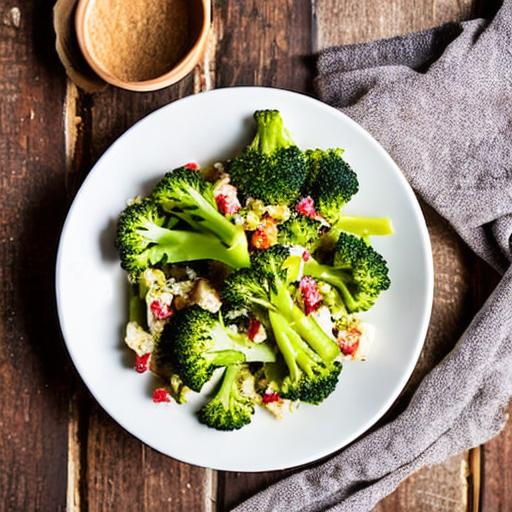} & 
\includegraphics[width=0.195\linewidth]{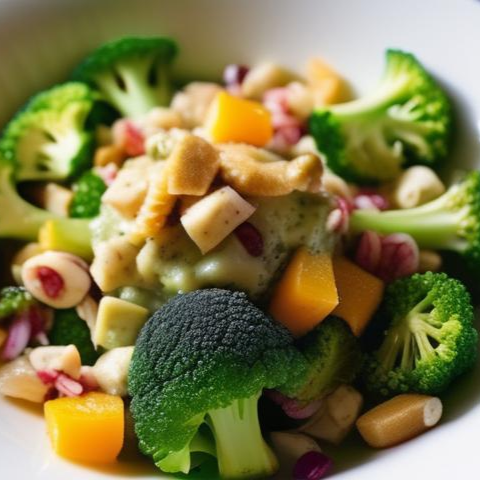} & & 
\includegraphics[width=0.195\linewidth]{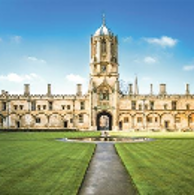} & 
\includegraphics[width=0.195\linewidth]{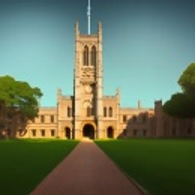} & 
\includegraphics[width=0.195\linewidth]{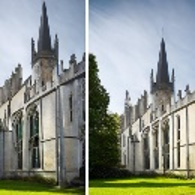} & 
\includegraphics[width=0.195\linewidth]{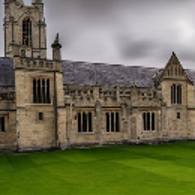} & 
\includegraphics[width=0.195\linewidth]{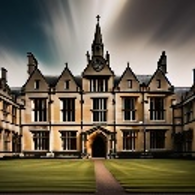} \\
\includegraphics[width=0.195\linewidth]{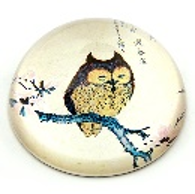} & 
\includegraphics[width=0.195\linewidth]{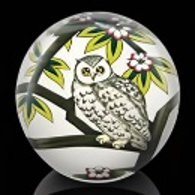} & 
\includegraphics[width=0.195\linewidth]{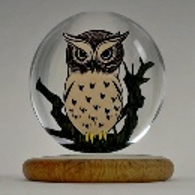} & 
\includegraphics[width=0.195\linewidth]{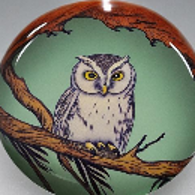} & 
\includegraphics[width=0.195\linewidth]{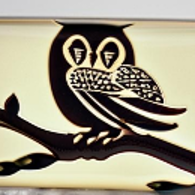} & & 
\includegraphics[width=0.195\linewidth]{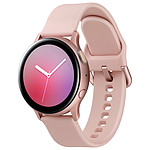} & 
\includegraphics[width=0.195\linewidth]{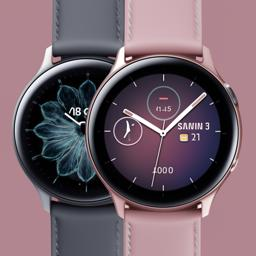} & 
\includegraphics[width=0.195\linewidth]{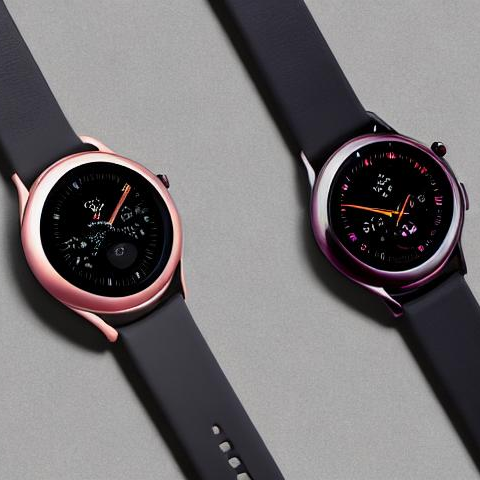} & 
\includegraphics[width=0.195\linewidth]{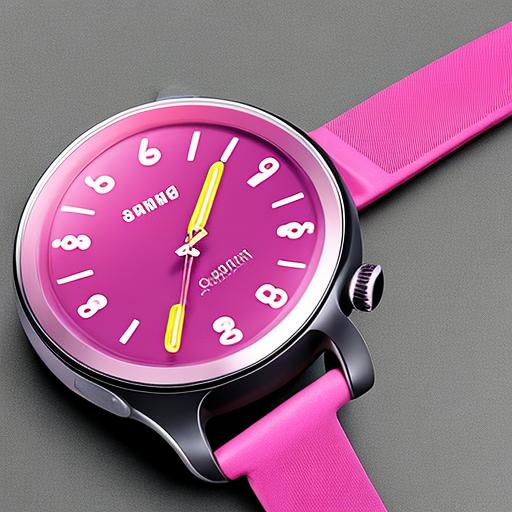} & 
\includegraphics[width=0.195\linewidth]{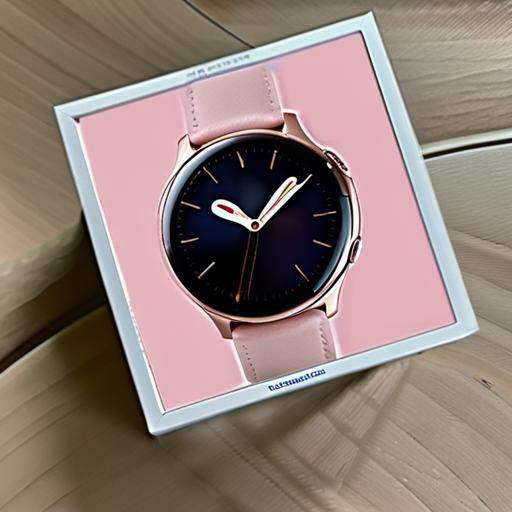} \\
\end{tabular}
}
\vspace{-0.3cm}
\caption{Qualitative samples from the proposed \dataset dataset. Each line considers a pristine image from LAION-400M~\cite{schuhmann2021laion} (left) and the four generated images (right).}
\label{fig:dataset}
\vspace{-0.4cm}
\end{figure}

As existing datasets are limited in their diversity of generators and quantity of images, we opt for creating and releasing a new dataset that can support learning an embedding space from scratch. Our \textbf{D}iffusion-generated \textbf{D}eepfake \textbf{D}etection dataset (D$^3$) contains nearly 2.3M records and 11.5M images. Each record in the dataset consists of a prompt, a real image, and four images generated with as many generators. Prompts and corresponding real images are taken from LAION-400M~\cite{schuhmann2021laion}, while fake images are generated, starting from the same prompt, using different text-to-image generators. Some sample records of our dataset are shown in Fig.~\ref{fig:dataset}.

\tit{Dataset collection}
We employ four state-of-the-art opensource diffusion models, namely Stable Diffusion v1.4 (SD-1.4)~\cite{rombach2022high}, Stable Diffusion v2.1 (SD-2.1)~\cite{rombach2022high}, Stable Diffusion XL (SD-XL)~\cite{podell2023sdxl}, and DeepFloyd IF (DF-IF)\footnote{\scriptsize\url{https://huggingface.co/DeepFloyd/IF-I-XL-v1.0}}. While the first three generators are variants of the Stable Diffusion approach, DeepFloyd IF is strongly inspired by Imagen~\cite{saharia2022photorealistic} and thus represents a different generation technique.

With the aim of increasing the variance of the dataset, images have been generated with different aspect ratios, \ie~$256^2$, $512^2$, 640$\times$480, and 640$\times$360. Moreover, to mimic the distribution of real images, we also employ a variety of encoding and compression methods (BMP, GIF, JPEG, TIFF, PNG). In particular, we closely follow the distribution of encoding methods of LAION itself, therefore favoring the presence of JPEG-encoded images.

We also adopt techniques to improve the quality of generated images from an aesthetic point of view. In particular, we employ negative prompts, which reduce the probability of generating selected concepts (\eg~blurry or low-quality images). Further, we employ common prompt engineering techniques for visual quality improvement, as well as modifiers. For safety reasons, prompts were filtered avoiding those tagged as not safe for work inside the dataset. We refer the reader to the supplementary for additional technical details about the generation.

\tit{Extended Test Set}
To increase the difficulty of the classification task and verify the generalization to unseen generators, we create an extended and more challenging test set by incorporating images generated by diverse state-of-the-art diffusion models.
Specifically, the considered unseen models encompass Stable Diffusion XL Turbo (SD-XL-T)~\cite{sauer2023adversarial},
Stable Diffusion unCLIP\footnote{\scriptsize\url{https://huggingface.co/stabilityai/stable-diffusion-2-1-unclip}}, Self-Attention Guidance (SAG)~\cite{hong2023improving},
aMUSEd~\cite{patil2024amused},
Kandinsky in its 2.1, 2.2, and 3 versions (K-2.1, K-2.2, K--3)~\cite{razzhigaev2023kandinsky}, and
PixArt-$\alpha$ (PA-$\alpha$)~\cite{chen2023pixart}. This test set also includes images generated by the four generators used to create the entire \dataset dataset. 

The generation always starts from textual prompts randomly extracted from the LAION-400M dataset, following the same procedure previously described. In this case, since the vast majority of images contained in the LAION-400M dataset are encoded in JPEG format, we maintain consistency by generating images in the same format. Moreover, to enhance reproducibility and ensure the possibility of directly releasing the real images contained in this split instead of image URLs, we select from the available LAION-400M images under CC-0 license. Additional information about this split can be found in the supplementary.

\tit{Comparison with existing datasets}
In Table~\ref{tab:dataset} we compare with existing datasets for deepfake detection. In particular, we consider the COCOFake~\cite{amoroso2023parents}, ELSA-1M\footnote{\scriptsize\label{elsa}\url{https://huggingface.co/datasets/elsaEU/ELSA1M\_track1}}, DiffusionDB~\cite{wang-etal-2023-diffusiondb}, and the dataset introduced in~\cite{ojha2023towards}, which all 
\begin{wraptable}{r}{0.66\textwidth}
\vspace{-0.9cm}
\centering
  \caption{Comparison between the \dataset dataset and previous datasets containing generated images.
  }
  \label{tab:dataset}
  \vspace{0.1cm}
  \centering
  \setlength{\tabcolsep}{.2em}
  \resizebox{0.98\linewidth}{!}{
  \begin{tabular}{lc cc c c ccc}
    \toprule
    & & & \multicolumn{2}{c}{\textbf{\#Gens}} \\
    \cmidrule{4-5}
   \textbf{Dataset} & \textbf{\#Ims} & & GANs & DMs & & \textbf{Public} & \textbf{Captions} & \textbf{Real Ims}  \\
    \midrule
    COCOFake~\cite{amoroso2023parents} & 720k & & - & 1 & & \cmark & \cmark & \cmark \\
    ELSA-1M\textsuperscript{\ref{elsa}} & 1M & & - & 1 & &  \cmark & \cmark & \cmark \\
    DiffusionDB~\cite{wang-etal-2023-diffusiondb} & 14M & & - & 1 & &  \cmark & \cmark & \xmark \\
    Simulacra AC\textsuperscript{\ref{simulacra}} & 240k & & 3 & - & &  \cmark & \cmark & \xmark \\
    CIFAKE~\cite{bird2023cifake} & 120k & & 1 & - & &  \cmark & \xmark & \cmark \\
    Wang~\etal~\cite{wang2020cnn} & 72k & & 11 & - & &  \cmark & \xmark & \cmark \\
    Ojha~\etal~\cite{ojha2023towards} & 800k & & - & 1 & &  \xmark & \xmark & \xmark \\  
    \midrule
    \rowcolor{teagreen}
    \textbf{\dataset (Ours)} & &  &  & &  & & & \\
    \rowcolor{teagreen}
    \hspace{0.3cm}Training Set & 12M & & - & 4 & &  \cmark & \cmark & \cmark \\
    \rowcolor{teagreen}
    \hspace{0.3cm}Test Set & 24k & & - & 4 & &  \cmark & \cmark & \cmark \\
    \rowcolor{teagreen}
    \hspace{0.3cm}Extended Test Set & 62k & & - & 12 & &  \cmark & \cmark & \cmark \\
    \bottomrule
  \end{tabular}
    }
  \vspace{-0.55cm}
\end{wraptable}
contain images generated from a single diffusion model. Further, we also consider datasets containing GAN images, such as Simulacra AC\footnote{\scriptsize\label{simulacra}\url{https://github.com/JD-P/simulacra-aesthetic-captions}}, CIFAKE~\cite{bird2023cifake}, and the one proposed by Wang~\etal~\cite{wang2020cnn}. Notably, \dataset is significantly larger than its existing counterparts, with the exception of DiffusionDB which, however, is not paired with real images and considers only one generator. Further, it is the only dataset containing more than one diffusion-based generator. As such, \dataset enables the development and testing of deepfake detection approaches that leverage paired real-fake samples and which are focused on multiple diffusion-based generators.

\section{Experiments\vspace{-0.1cm}}
\label{sec:experiments}
\subsection{Experimental Setting\vspace{-0.01cm}}
\label{sec:expsettings}
\tinytit{Implementation and training details}
We employ the Tiny version of the standard Vision Transformer architecture~\cite{touvron2021training} as our backbone $\mathcal{F}$, and train it from scratch. Given that our approach relies on a contrastive learning training objective, the use of a smaller model allows us to increase the batch size, which is configured to $256$ for each GPU. We set the learning rate at 2$e^{-3}$ and employ the AdamW optimizer~\cite{DBLP:conf/iclr/LoshchilovH19}. During training, the number of image transformations applied through the $\mathcal{T}$ operator is uniformly chosen between 0 and 2 for each image. After eventually applying random transformations, all images are randomly cropped to $224^2$. When computing the final loss, we apply a constant weight of 1.0 to both loss components, $\mathcal{L}_\text{global}$ and $\mathcal{L}_\text{multi-scale}$. Overall, our training takes two days with 4 GPUs.
We refer the reader to the supplementary material for detailed training and transformations hyper-parameters. 
\tit{Classifiers training details} To train the linear, nearest neighbor, and one-class SVM classifiers, a set of 9,600 records is randomly sampled from the training split and is then randomly transformed. Following this, we gather features by applying the backbone to the selected real and fake images. The nearest neighbor classifier subsequently utilizes these feature banks as a repository for making real-fake predictions. For the linear classifier, we employ a maximum of 1,500 training iterations and a balanced loss function that accounts for the discrepancy in the frequency of real and fake samples. This is crucial, as the prevalence of fake data is fourfold compared to real data within the training collection.
When fitting the one-class SVM classifier~\cite{NIPS1999_8725fb77}, we only utilize the real images contained in the 9,600 selected records. 
To create a correct boundary among different sources of images, we consider a polynomial kernel with the $\nu$ parameter set to $0.1$.
\tit{Evaluation} When testing on \dataset, we consider both a standard test set containing images generated using the four diffusion models contained in the training split. Under this setting, we consider both a case in which test images remain unaltered and a case where test images are randomly transformed. Additionally, we report experiments on the extended test set described in Sec.~\ref{sec:dataset}. Both test splits are composed of 4,800 different records.

\subsection{Experimental Results}
\tinytit{Evaluation on D$^3$ dataset}
To validate our proposal, we first conduct experiments on the \dataset dataset. As previously mentioned, our comparisons encompass various pre-trained visual backbones followed by a linear, a nearest neighbor (NN), or a one-class SVM classifier fitted on the same 9,600 records from \dataset used in our model. In particular, we use a standard ViT Tiny (ViT-T)~\cite{touvron2021training} pre-trained on ImageNet-21k~\cite{5206848}, a CLIP-based model in its ViT Base (ViT-B) and ViT Large (ViT-L) versions~\cite{radford2021learning}, and a backbone pre-trained with self-supervised objectives like DINOv2~\cite{oquab2023dinov2}, using the ViT Base model. To verify the effectiveness of using contrastive learning, we also train a ViT-Tiny model from scratch on our dataset with a binary cross-entropy (BCE) loss.

In addition to these models, we also compare with existing approaches for deepfake detection specifically tailored to recognize fake images from GAN generators. Specifically, we include the models proposed by Wang~\etal~\cite{wang2020cnn} which are based on a ResNet-50 (RN50)~\cite{he2016identity} model trained with different image transformations. Other considered approaches involve variations of the ResNet-50 architecture specifically trained with images sourced from GANs~\cite{gragnaniello2021gan} and from latent diffusion models~\cite{corvi2023detection}.
Furthermore, we consider the approach proposed by Wang~\etal~\cite{wang2023dire} that takes into account the reconstructed image and its disparities with the original counterpart and the model presented by Ojha~\etal~\cite{ojha2023towards} which leverages a CLIP model based on the ViT Large architecture, followed by a linear classifier. Noteworthy, \ours employs 5M parameters compared to RN50 (26M), ViT-B (86M), and ViT-L (307M) resulting in a smaller and more efficient model.
For all competitors, we use the pre-trained weights downloaded from the official repositories provided by the authors.

\begin{table*}[t]
  \caption{Performance evaluation on \dataset test set considering settings with and without image transformations. ``Overall'' accuracy averages predictions for both fake and real images. ``Fake'' accuracy denotes the mean accuracy for generated data.
  The $\dagger$ marker indicates methods trained on other datasets and tested on the \dataset test set.
  }
  \label{tab:all_generators}
  \vspace{-0.15cm}
  \centering
  \setlength{\tabcolsep}{.3em}
  \resizebox{\linewidth}{!}{
  \begin{tabular}{lc ccc cc ccc c cccc}
    \toprule
    & & \multicolumn{3}{c}{\textbf{w/o Transforms}} & & & \multicolumn{8}{c}{\textbf{w/ Transforms}} \\
    \cmidrule{3-5} \cmidrule{8-15}
    \textbf{Model} & & Overall & Real & Fake & & & Overall & Real & Fake & & DF-IF & SD-1.4 & SD-2.1 & SD-XL \\
    \midrule
    Wang~\etal~(RN50 Blur+JPEG 0.5)~\cite{wang2020cnn}$^\dagger$ & & 20.7 & \underline{99.4} & 1.0 & & & 20.8 & \underline{99.2} & 1.2 & & 0.9 & 1.6 & 1.2 & 1.4 \\
    Wang~\etal~(RN50 Blur+JPEG 0.1)~\cite{wang2020cnn}$^\dagger$ & & 21.4 & 98.7 & 2.0 & & & 21.6 & 98.2 & 2.5 & & 2.2 & 2.8 & 2.1 & 2.8 \\
    Gragnaniello~\etal~\cite{gragnaniello2021gan}$^\dagger$ & & 21.8 & \textbf{99.7} & 2.3 & & & 21.8 & \textbf{99.5} & 2.3 & & 1.4 & 4.2 & 1.5 & 2.1 \\
    Corvi~\etal~\cite{corvi2023detection}$^\dagger$  & & 75.9 & 99.2 & 70.1 & & & 64.1 & \underline{99.2} & 55.4 & & 8.1 & 84.1 & 76.0 & 53.3 \\
    Ojha~\etal~\cite{ojha2023towards}$^\dagger$ & & 31.0 & 96.1 & 14.8 & & & 37.7 & 87.0 & 25.4 & & 11.3 & 24.5 & 19.0 & 46.8 \\
    Wang~\etal~(DIRE)~\cite{wang2023dire}$^\dagger$ & & 79.7 & 10.0 & 97.1 & & & 76.5 & 15.8 & 91.7 & & 89.6 & 92.4 & 91.5 & 93.1 \\
    \midrule
    ViT-T (Linear) & & 88.2 & 85.8 & 88.8 & & & 81.4 & 81.0 & 81.5 & & 76.7 & 76.6 & 81.6 & 90.9 \\
    CLIP ViT-B (Linear) & & 89.1 & 81.0 & 91.1 & & & 87.4 & 86.8 & 87.5 & & 80.4 & 87.0 & 87.7 & 95.2 \\
    DINOv2 ViT-B (Linear) & & 85.5 & 81.8 & 86.4 & & & 83.0 & 82.0 & 83.2 & & 70.6 & 81.8 & 85.5 & 94.9 \\
    CLIP ViT-L (Linear) & & 92.5 & 88.3 & 93.5 & & & 89.3 & 89.0 & 89.4 & & 83.5 & 88.9 & 90.2 & 95.0 \\
    \midrule
    ViT-T (NN) & & 81.8 & 40.0 & 92.2 & & & 80.9 & 40.7 & 90.9 & & 88.8 & 89.4 & 90.9 & 94.9 \\
    CLIP ViT-B (NN) & & 82.8 & 38.7 & 93.8 & & & 81.6 & 38.9 & 92.3 & & 87.6 & 92.5 & 93.5 & 95.7 \\
    DINOv2 ViT-B (NN) & & 79.2 & 37.5 & 89.6 & & & 87.1 & 78.9 & 89.2 & & 84.9 & 88.2 & 90.0 & 93.7 \\
    CLIP ViT-L (NN) & & 83.4 & 40.6 & 94.0 & & & 82.1 & 38.9 & 92.9 & & 87.7 & 94.0 & 94.0 & 96.0 \\
    CLIP ViT-L (SVM) & & 23.5 & 86.8 & 7.7 & & & 22.6 & 90.3 & 5.7 & & 6.3 & 4.8 & 4.7 & 7.1 \\
    \midrule
     ViT-T (BCE) & & 97.0 & 91.4 & 98.4 & & & 93.7 & 93.8 & 93.6 & & 92.1 & 93.5 & 92.7 & 96.4 \\
    \rowcolor{teagreen}
    \textbf{\ours (Linear)} & & \textbf{98.0} & 94.0 & \underline{99.0} & & & \underline{95.7} & 95.6 & \underline{95.8} & & \underline{94.8} & \underline{95.8} & \underline{94.9} & 97.5 \\
    \rowcolor{teagreen}
    \textbf{\ours (NN)} & & \underline{97.3} & 89.3 & \textbf{99.3} & & & \textbf{95.8} & 90.5 & \textbf{97.1} & & \textbf{96.6} & \textbf{97.3} & \textbf{96.6} & \underline{98.1} \\
    \rowcolor{teagreen}
    \textbf{\ours (SVM)} & & 91.2 & 74.4 & 95.4 & & & 92.5 & 81.0 & 95.4 & & \textbf{96.6} & 91.7 & 94.4 & \textbf{99.0} \\
    \bottomrule
  \end{tabular}
}
\vspace{-0.5cm}
\end{table*}

In Table~\ref{tab:all_generators}, we show the results obtained by the \ours approach and all the previously introduced deepfake detectors. Results are reported on the standard test split of the \dataset dataset with and without image transformations. As it can be seen, deepfake detectors trained on fake images generated by GANs face difficulties in generalizing to images generated by diffusion models within the D$^3$ dataset. For instance, both the approach proposed by Wang~\etal~\cite{wang2020cnn} and the one proposed in~\cite{gragnaniello2021gan} achieve a mean accuracy of less than $3\%$ across all fake generators. The CLIP-based ViT-L model followed by a linear classifier fitted on GAN-generated images~\cite{ojha2023towards} surpasses other GAN-based detectors with an overall accuracy of $31.0\%$, but still averages $14.8\%$ accuracy on all fake images, falling below random choice probability. Differently, the model introduced by Corvi~\etal~\cite{corvi2023detection}, which is trained on fake images generated from Latent Diffusion~\cite{rombach2022high}, obtains 70.1\% of accuracy on average on fake data when tested without transformation. Conversely, the accuracy drops to 55.4\% when transformations are applied showcasing sensitivity to underline augmentation. While DIRE~\cite{wang2023dire} obtains an average of 97.1\% on fake data, performance on real data reduces to 15.8\% and 10\% respectively in the splits with and without transformations.  

Notably, our results demonstrate that \ours outperforms all the pre-trained models with all considered classifiers in both settings. Specifically, when considering the setting without image transformations, \ours improves the final performance by 5.5\% and 4.8\% in terms of overall accuracy compared to the best-performing competitor (\ie~CLIP ViT-L with the linear classifier fitted on our data), respectively when using linear and nearest neighbor classifiers. When instead considering the setting with image transformations, the performance gain achieved by the best version of \ours is significantly higher, with an improvement of 6.5\% compared to the CLIP ViT-L model with a linear classifier, highlighting the appropriateness of employing global-local correspondences together with a contrastive objective. This is confirmed also when comparing our results with those obtained by the ViT-T model trained with BCE loss, which achieves competitive but lower performance on both considered settings. 

It is also important to note that applying a one-class SVM classifier to a pre-trained backbone like CLIP ViT-L leads to very poor results with an overall accuracy slightly above 20\%. This can be explained by the unsupervised nature of the employed SVM classifier, which is fitted only considering real images. Instead, using the one-class SVM along with our backbone can significantly increase the results compared to CLIP ViT-L with the same classifier, further demonstrating the goodness of our embedding space.

\begin{table*}[t]
  \caption{Performance evaluation on \dataset extended test set. We report accuracy scores over real images and fake ones from each of the 12 generators included in the test set. Additionally, we include the average accuracy across all fake images (Avg) and those generated by diffusion models not seen during training (Avg$_\text{u}$). The $\dagger$ marker indicates methods trained on other datasets and tested on the \dataset extended test set.}
  \label{tab:extended_test_set}
  \vspace{-0.15cm}
  \centering
  \setlength{\tabcolsep}{.18em}
  \resizebox{\linewidth}{!}{
  \begin{tabular}{lc cc cccccccccccc cc}
    \toprule
    \textbf{Model} & & Real & & DF-IF & SD-1.4 & SD-2.1 & SD-XL & SD-XL-T & unCLIP & SAG & aMUSEd & K-2.1 & K-2.2 & K-3 & PA-$\alpha$ & \textbf{Avg} & \textbf{Avg$_\text{u}$} \\
    \midrule
    Wang~\etal~(RN50 0.5)~\cite{wang2020cnn}$^\dagger$ & &  99.2 & & 0.3 & 1.2 & 0.7 & 2.0 & 5.5 & 2.3 & 0.9 & 0.9 & 0.4 & 0.5 & 0.8 & 0.4 & 1.3 & 1.5 \\
    Wang~\etal~(RN50 0.1)~\cite{wang2020cnn}$^\dagger$ & & 97.9 & & 0.9 & 1.2 & 0.6 & 4.2 & 5.4 & 3.4 & 1.0 & 1.2 & 1.1 & 1.1 & 1.4 & 0.6 & 1.8 & 1.9 \\
    Gragnaniello~\etal~\cite{gragnaniello2021gan}$^\dagger$ & & \textbf{99.6} & & 0.3 & 2.7 & 0.4 & 0.3 & 0.2 & 11.9 & 3.4 & 1.6 & 0.4 & 0.1 & 0.2 & 04. & 1.8 & 2.3 \\
    Corvi~\etal~\cite{corvi2023detection}$^\dagger$ & & \underline{99.4} & & 6.4 & \textbf{99.9} & \underline{98.9} & 71.4 & 90.8 & \textbf{99.4} & \underline{99.4} & 78.5 & 91.5 & 89.2 & 58.8 & \textbf{98.5} & 81.9 & 88.2 \\
    Ojha~\etal~\cite{ojha2023towards}$^\dagger$ & & 95.9 & & 1.7 & 11.0 & 7.7 & 14.8 & 47.0 & 53.3 & 9.2 & 40.4 & 15.0 & 8.8 & 14.2 & 14.2 & 19.8 & 25.3 \\
    \midrule
    ViT-T (Linear) & & 53.9 & & 89.5 & 82.9 & 87.9 & 78.4 & 84.9 & 81.7 & 84.9 & 87.3 & 79.6 & 86.5 & 80.2 & 85.8 & 84.1 & 83.9 \\
    ViT-T (NN) & & 24.9 & & 90.8 & 90.7 & 91.9 & 89.3 & 88.0 & 87.1 & 91.5 & 91.6 & 88.5 & 90.3 & 89.2 & 90.2 & 89.9 & 89.5 \\
    DINOv2 ViT-B (Linear) & & 64.7 & & 69.9 & 81.4 & 80.3 & 75.9 & 80.8 & 71.7 & 83.3 & 92.9 & 77.8 & 84.6 & 76.2 & 84.9 & 80.0 & 81.5 \\
    DINOv2 ViT-B (NN) & & 22.5 & & 86.0 & 89.1 & 89.4 & 87.3 & 84.3 & 83.9 & 89.0 & 90.4 & 87.3 & 89.0 & 85.8 & 89.1 & 87.6 & 87.4 \\
    CLIP ViT-L (Linear) & & 56.3 & & \underline{96.4} & 98.5 & 98.3 & 79.0 & 91.0 & 88.5 & 97.9 & \textbf{99.8} & 82.6 & 90.4 & 81.6 & 89.1 & 91.1 & 90.1 \\
    CLIP ViT-L (NN) & & 16.8 & & 94.0 & 96.9 & 97.0 & 93.5 & \underline{94.1} & 95.6 & 96.6 & \underline{96.5} & \textbf{92.8} & \underline{93.9} & \textbf{92.4} & 94.0 & \underline{94.7} & \underline{94.4} \\
    \midrule
    ViT-T (BCE) & & 92.6 & & 85.5 & 97.8 & 98.2 & 93.9 & 87.8 & 95.2 & 99.0 & 79.5 & 73.1 & 83.6 & 68.5 & 84.6 & 87.2 & 83.9 \\
    \rowcolor{teagreen}
    \textbf{\ours (Linear)} & & 97.3 & & 85.2 & 98.8 & 98.7 & 92.9 & 86.4 & 95.0 & 99.2 & 72.8 & 70.5 & 82.8 & 67.1 & 86.5 & 86.3 & 82.5 \\
    \rowcolor{teagreen}
    \textbf{\ours (NN)} & & 95.3 & & 90.2 & \underline{99.3} & \textbf{99.2} & \underline{95.1} & 91.2 & \underline{97.2} & \textbf{99.5} & 81.4 & 78.5 & 87.5 & 75.8 & 91.1 & 90.5 & 87.8 \\
     \rowcolor{teagreen}
    \textbf{\ours (SVM)} & & 91.9 & & \textbf{96.9} & 91.0 & 94.8 & \textbf{98.6} & \textbf{97.0} & 93.0 & 95.1 & 94.4 & \underline{92.2} & \textbf{97.5} & \underline{92.2} & \underline{96.2} & \textbf{94.8} & \textbf{94.7} \\
    \bottomrule
  \end{tabular}
}
\vspace{-0.5cm}
\end{table*}

\tit{Generalization performance on unseen generators}
In the realm of deepfake detection, the ability to generalize to various types of generators not encountered during training is crucial for readiness in real-world scenarios. We therefore consider a more challenging setting with fake images generated by diverse diffusion-based models not seen during training.  In Table~\ref{tab:extended_test_set}, we report the performance of \ours and competitors on the extended test set of \dataset. Here, in addition to providing accuracy results over real images and fake ones from each of the 12 considered generators, we include accuracy scores averaged across all fake images and only those generated by diffusion models not seen during training (\ie~across fake images generated by 8 different diffusion models). 

\begin{table}[t]
  \caption{Accuracy results on external unseen generators employing data from a mixture of diffusion and autoregressive models. Following~\cite{ojha2023towards}, we average the accuracy scores between real and fake images. The $\dagger$ marker indicates methods trained on datasets other than \dataset.}
  \label{tab:datasets_diffusion}
  \vspace{-0.15cm}
  \centering
  \setlength{\tabcolsep}{.25em}
  \resizebox{\linewidth}{!}{
  \begin{tabular}{lc c c ccc c ccc c ccc c c c c}
    \toprule
    & & & & \multicolumn{3}{c}{LDM} & & \multicolumn{3}{c}{GLIDE} & & \multicolumn{3}{c}{DALL-E} \\
    \cmidrule{5-7} \cmidrule{9-11} \cmidrule{13-15}
   \textbf{Model} & & Guided & & 200 & 200 (CFG) & 100 & & 100 (27) & 50 (27) & 100 (10) & & v1 & v2 & v3 & & Midjourney & & \textbf{Avg} \\
    \midrule
    Wang~\etal~(RN50 0.5)~\cite{wang2020cnn}$^\dagger$ & & 52.0 & & 51.1 & 51.4 & 51.3 & & 53.3 & 55.6 & 54.3 & & 52.5 & 50.9 & 49.8 & & 50.1 & & 52.4 \\
    Wang~\etal~(RN50 0.1)~\cite{wang2020cnn}$^\dagger$ & & \underline{62.0} & & 53.9 & 55.3 & 55.1 & & 60.3 & 62.7 & 61.0 & & 56.1 & 66.2 & 50.2 & & 52.2 & & 57.7 \\
    Gragnaniello~\etal~\cite{gragnaniello2021gan}$^\dagger$ & & 54.1 & & 58.0 & 61.1 & 57.5 & & 56.9 & 59.6 & 58.8 & & 71.7 & 57.1 & 50.1 & & 50.9 & & 57.8 \\
    Corvi~\etal~\cite{corvi2023detection}$^\dagger$ & & 52.1 & & \textbf{99.3} & \textbf{99.3} & \textbf{99.3} & & 58.0 & 59.1 & 62.3 & & \textbf{89.4} & 49.6 & 82.9 & & \textbf{98.3} & & 77.2 \\
    Ojha~\etal~\cite{ojha2023towards}$^\dagger$ & & \textbf{69.5} & & \underline{94.4} & 74.0 & \underline{95.0} & & \underline{78.5} & \underline{79.1} & \underline{77.9} & & \underline{87.3} & 60.1 & 53.5 & & 53.9 & & 74.8 \\
    Wang~\etal~(DIRE)~\cite{wang2023dire}$^\dagger$ & & 56.7 & & 62.6 & 61.3 & 62.2 & & 63.2 & 63.4 & 63.1 & & 63.0 & 63.4 & 60.7 & & 62.2 & & 62.0 \\
    \midrule
    ViT-T (Linear) & & 46.7 & & 52.2 & 60.7 & 53.9 & & 55.7 & 54.6 & 57.5 & & 71.3 & \underline{87.3} & 83.8 & & 78.1 & & 63.8 \\
    ViT-T (NN) & & 48.4 & & 56.0 & 58.4 & 55.6 & & 56.3 & 53.7 & 56.4 & & 60.4 & 64.9 & 64.7 & & 63.4 & & 58.0 \\
    DINOv2 ViT-B (Linear) & & 55.1 & & 72.6 & 72.2 & 74.1 & & \textbf{81.1} & \textbf{79.7} & \textbf{83.5} & & 72.4 & \textbf{93.7} & 81.5 & & 81.4 & & 77.0 \\
    DINOv2 ViT-B (NN) & & 50.2 & & 61.0 & 60.1 & 60.1 & & 60.7 & 60.3 & 60.6 & & 60.4 & 64.4 & 62.6 & & 62.9 & & 60.3 \\
    CLIP ViT-L (Linear) & & 53.0 & & 78.0 & 73.6 & 77.4 & & 74.5 & 76.7 & 77.2 & & 82.7 & \underline{87.3} & 87.2 & & 84.8 & & 77.5 \\
    CLIP ViT-L (NN) & & 52.1 & & 66.1 & 62.0 & 64.9 & & 62.6 & 62.5 & 62.8 & & 64.3 & 67.8 & 67.8 & & 67.1 & & 63.6 \\
    \midrule
    ViT-T (BCE) & & 52.9 & & 82.8 & 90.7 & 82.1 & & 72.6 & 75.8 & 73.6 & & 62.8 & 71.5 & 85.6 & & 76.4 & & 75.1 \\
    \rowcolor{teagreen}
    \textbf{\ours (Linear)} & & 53.5 & & 92.5 & 95.6 & 91.9 & & 71.7 & 75.4 & 72.9 & & 63.1 & 71.4 & 86.7 & & 84.0 & & 78.0 \\
    \rowcolor{teagreen}
    \textbf{\ours (NN)} & & 53.5 & & 92.7 & \underline{96.1} & 92.5 & & 73.8 & 76.9 & 74.0 & & 67.0 & 74.3 & \underline{88.6} & & 86.8 & & \underline{79.6} \\
    \rowcolor{teagreen}
    \textbf{\ours (SVM)} & & 54.6 & & 91.0 & 90.4 & 90.9 & & 77.2 & 78.8 & 77.6 & & 76.1 & 80.2 & \textbf{91.0} & & \underline{89.7} & & \textbf{81.6} \\
    \bottomrule
  \end{tabular}
}
\vspace{-0.5cm}
\end{table}

As previously highlighted, detectors trained on GAN-generated images~\cite{wang2020cnn,gragnaniello2021gan} exhibit sub-optimal performance on this collection, misclassifying synthetically generated images as authentic data. This phenomenon is reflected in the near-perfect accuracy rates exceeding 97\% for authentic images, against an average accuracy not exceeding 2\% on fake data for all the models. Conversely, the pre-trained models equipped with both linear and nearest neighbor classifiers tend to underperform on the real distribution. For instance, CLIP ViT-L and ViT-T equipped with linear classifiers respectively achieve only 56.3\% and 53.9\% of accuracy on real data. On the other hand, \ours with one-class SVM achieves the best results on unseen generators on average with an accuracy score of 94.7\% with a gain of 6.5\% and 69.4\% respectively compared to the models proposed in~\cite{corvi2023detection} and~\cite{ojha2023towards}. Noteworthy, both versions of \ours with nearest neighbor and one-class SVM classifier outperform the baseline ViT-T trained with BCE loss, thus validating the relevance of our training protocol also when recognizing fake images generated by unseen generators. 

In addition to the evaluation on our extended test set, we compare \ours with other deepfake detectors following the datasets adopted by~\cite{ojha2023towards}. 
Specifically, we perform experiments using the images generated by  Guided~\cite{dhariwal2021diffusion}, LDM~\cite{rombach2022high}, GLIDE~\cite{nichol2021glide}, DALL-E~\cite{ramesh2021zero}. Additionally, we collect images from diverse non-public and commercial generative tools like DALL-E 2~\cite{ramesh2022hierarchical}, DALL-E 3~\cite{betker2023improving}, and Midjourney V5. 
Following previous literature, in this setting we employ 1,000 real images and 1,000 fake ones for each considered generator and report the average accuracy over both real and fake data. Images from Guided~\cite{dhariwal2021diffusion} are paired with 1,000 real images from ImageNet, while all other fake sources are paired with real images from the LAION split provided by~\cite{ojha2023towards}.
Results are shown in Table~\ref{tab:datasets_diffusion}. As it can be seen, also in this setting \ours demonstrates superior performance than competitors with an overall average accuracy of 81.6\% using the one-class SVM classifier. Notably, the best deepfake recognition results are achieved on more recent diffusion-based models like DALL-E 2, DALL-E 3, and Midjourney V5 which generally generates highly realistic images. 

\tit{Ablation study on different contrastive pre-trainings}
As detailed in Sec.~\ref{sec:method}, the \ours model is trained using a combination of two loss functions (\ie~$\mathcal{L}_\text{global}$ and $\mathcal{L}_\text{multi-scale}$). The first loss operates on entire images while the second operates on local and global crops. 
In Table~\ref{tab:ablation}, we report an ablation study on the key components of our approach. 
Firstly, we consider \ours trained without applying the transformation pipeline. Secondly, we compare our complete model with a model trained using only the $\mathcal{L}_\text{global}$ trained from scratch or starting from the weights of the ViT Tiny model pre-trained on ImageNet-21k using the entire loss formulation. Further, we consider an additional backbone trained only with a contrastive loss between real and fake. Lastly, we analyze the performance of models trained with $\mathcal{L}_\text{multi-scale}$ that leverages only local or global crops which differs from \ours that employs both. 

As it can be seen, \ours achieves the best results on average on the \dataset test set with transformations. Differently, training the model with $\mathcal{L}_\text{global}$ (real $\rightarrow$ fake) or without applying transformations leads to the worst results in this setting.
Additionally, fine-tuning a pre-trained ViT Tiny model with $\mathcal{L}_{\text{global}}$ does not yield improvements in terms of detection accuracy. This confirms that training the backbone from scratch is beneficial for the final performance. Overall, \ours outperforms other components on the split with transformations, improving the performance by $3.2\%$, $0.9\%$, and $0.3\%$ respectively when fitting a linear, NN, and one-class SVM classifier in comparison to the model variant with $\mathcal{L}_\text{global}$ only. Performance increases by $0.7\%$, $0.5\%$, and $3.8\%$ when comparing \ours with the version trained with only local crops, and comparable gains are obtained when employing global crops. This result proves the utility of the $\mathcal{L}_\text{multi-scale}$ loss function, underlying the relevance of the alignment of local-global views.

\begin{table}[t]
  \caption{
  Contribution of each loss component within our proposed methodologies. The accuracy is computed on the \dataset test set with and without image transformations.
  }
  \label{tab:ablation}
    \vspace{-0.15cm}
  \centering
  \setlength{\tabcolsep}{.3em}
  \resizebox{0.95\linewidth}{!}{
  \begin{tabular}{lc ccc cc ccc c cccc}
    \toprule
    & & \multicolumn{3}{c}{\textbf{w/o Transforms}} & & & \multicolumn{8}{c}{\textbf{w/ Transforms}} \\
    \cmidrule{3-5} \cmidrule{8-15}
    \textbf{Model} & & Overall & Real & Fake & & & Overall & Real & Fake & & DF-IF & SD-1.4 & SD-2.1 & SD-XL \\
    \midrule
    w/o transforms & & 96.4 & 86.2 & 98.9 & & & 87.2 & 89.7 & 86.6 & & 83.4 & 86.6 & 85.3 & 91.4 \\
    w/ $\mathcal{L}_\text{global}$ only & & 97.7 & 93.5 & 98.8 & & & 94.5 & 95.2 & 94.4 & & 92.8 & 94.6 & 93.5 & 96.7 \\
    w/ $\mathcal{L}_\text{global}$ only (pre-trained) & & 87.3 & 93.8 & 85.7 & & & 86.2 & 92.9 & 84.5 & & 94.0 & 76.6 & 76.7 & 91.0 \\
    w/ $\mathcal{L}_\text{global}$ only (real $\rightarrow$ fake) & & 87.7 & 74.9 & 90.9 & & & 83.5 & 75.3 & 85.6 & & 80.7 & 84.6 & 86.5 & 90.5 \\
    only local crops & & 97.5 & 92.5 & 98.8 & & & 94.8 & 95.1 & 94.8 & & 93.7 & 94.8 & 93.9 & 96.9 \\
    only global crops & & 97.6 & 93.7 & 98.6 & & & 95.0 & 94.9 & 95.1 & & 93.3 & 94.2 & \textbf{96.3} & 96.5 \\
    \rowcolor{teagreen}
    \textbf{\ours (Linear)} & & \textbf{98.0} & \textbf{94.0} & \textbf{99.0} & & & \textbf{95.7} & \textbf{95.6} & \textbf{95.8} & & \textbf{94.8} & \textbf{95.8} & 94.9 & \textbf{97.5} \\
    \midrule
    w/o transforms & & 93.5 & 69.9 & \textbf{99.4} & & & 88.6 & 74.7 & 92.0 & & 90.0 & 92.2 & 91.6 & 94.3 \\
    w/ $\mathcal{L}_\text{global}$ only & & \textbf{97.4} & 91.0 & 99.0 & & & 94.9 & 91.5 & 95.9 & & 94.8 & 95.6 & 95.6 & 97.5 \\
    w/ $\mathcal{L}_\text{global}$ only (pre-trained) & & 96.1 & 86.8 & 98.4 & & & 94.2 & 86.5 & 96.2 & & 95.3 & 96.0 & 95.4 & 98.0 \\
    w/ $\mathcal{L}_\text{global}$ only (real $\rightarrow$ fake) & & 76.2 & 75.1 & 76.5 & & & 73.9 & 75.3 & 73.5 & & 68.3 & 70.3 & 73.9 & 81.5 \\
    only local crops & & 97.1 & 89.3 & 99.0 & & & 95.3 & 91.3 & 96.3 & & 94.9 & 96.5 & 95.9 & 97.8 \\
    only global crops & & \textbf{97.4} & \textbf{91.6} & 98.8 & & & 94.3 & \textbf{92.0} & 94.9 & & 93.8 & 95.1 & 94.1 & 96.6 \\
    \rowcolor{teagreen}
    \textbf{\ours (NN)} & & 97.3 & 89.3 & 99.3 & & & \textbf{95.8} & 90.5 & \textbf{97.1} & & \textbf{96.6} & \textbf{97.3} & \textbf{96.6} & \textbf{98.1} \\
    \midrule
    w/o transforms & & 95.1 & 75.8 & \textbf{99.9} & & & 84.9 & \textbf{87.9} & 84.2 & & 79.7 & 85.4 & 84.2 & 87.6 \\
    w/ $\mathcal{L}_\text{global}$ only & & \textbf{95.8} & 80.2 & 99.7 & & & 92.2 & 87.8 & 93.4 & & 93.5 & 92.8 & 91.9 & 95.3 \\
    w/ $\mathcal{L}_\text{global}$ only (pre-trained) & & 89.2 & 46.4 & \textbf{99.9} & & & 89.1 & 46.0 & \textbf{99.9} & & \textbf{99.9} & \textbf{99.9} & \textbf{99.9} & \textbf{99.9} \\
    w/ $\mathcal{L}_\text{global}$ only (real $\rightarrow$ fake) & & 80.1 & \textbf{86.7} & 78.5 & & & 74.9 & 86.2 & 72.1 & & 66.2 & 67.8 & 71.2 & 83.4 \\
    only local crops & & 91.4 & 67.2 & 97.5 & & & 88.7 & 73.1 & 92.6 & & 97.5 & 93.0 & 92.5 & 87.5 \\
    only global crops & & 88.5 & 71.1 & 92.7 & & & 86.4 & 76.0 & 89.0 & & 86.0 & 91.3 & 92.8 & 85.9 \\
    \rowcolor{teagreen}
    \textbf{\ours (SVM)} & & 91.2 & 74.4 & 95.4 & & & \textbf{92.5} & 81.0 & 95.4 & & 96.6 & 91.7 & 94.4 & 99.0 \\
    \bottomrule
  \end{tabular}
}
\vspace{-0.5cm}
\end{table}

\section{\vspace{-0.05cm}Conclusion\vspace{-0.1cm}}
\label{sec:conclusion}
We introduced \ours, a methodology aimed at cultivating a contrastive-learned embedding space designed for the purpose of detecting deepfake content. Our method strategically incorporates a global contrastive loss, simultaneously emphasizing global-local correspondences via local and global crops. The training of our model involved the creation of the \dataset dataset, comprising 9.2 million images generated through the utilization of a large variety of state-of-the-art diffusion models. Experimental results demonstrate the superior performance and enhanced generalization capabilities exhibited by our proposed approach.

\section*{Acknowledgments} We acknowledge the CINECA award under the ISCRA initiative, for the availability of high-performance computing resources and support. This work has been supported by the Horizon Europe project ``European Lighthouse on Safe and Secure AI (ELSA)'' (HORIZON-CL4-2021-HUMAN-01-03), co-funded by the European Union.

%
%
\bibliographystyle{splncs04}
\bibliography{main}

\begin{thebibliography}{10}
\providecommand{\url}[1]{\texttt{#1}}
\providecommand{\urlprefix}{URL }
\providecommand{\doi}[1]{https://doi.org/#1}

\bibitem{amoroso2023parents}
Amoroso, R., Morelli, D., Cornia, M., Baraldi, L., Del~Bimbo, A., Cucchiara, R.: {Parents and Children: Distinguishing Multimodal DeepFakes from Natural Images}. ACM TOMM  (2024)

\bibitem{balaji2022ediffi}
Balaji, Y., Nah, S., Huang, X., Vahdat, A., Song, J., Kreis, K., Aittala, M., Aila, T., Laine, S., Catanzaro, B., et~al.: {eDiff-I: Text-to-Image Diffusion Models with an Ensemble of Expert Denoisers}. arXiv preprint arXiv:2211.01324  (2022)

\bibitem{betker2023improving}
Betker, J., Goh, G., Jing, L., Brooks, T., Wang, J., Li, L., Ouyang, L., Zhuang, J., Lee, J., Guo, Y., et~al.: Improving image generation with better captions (2023)

\bibitem{bird2023cifake}
Bird, J.J., Lotfi, A.: {CIFAKE: Image Classification and Explainable Identification of AI-Generated Synthetic Images}. arXiv preprint arXiv:2303.14126  (2023)

\bibitem{brock2018large}
Brock, A., Donahue, J., Simonyan, K.: {Large Scale GAN Training for High Fidelity Natural Image Synthesis}. ICLR  (2019)

\bibitem{caron2020unsupervised}
Caron, M., Misra, I., Mairal, J., Goyal, P., Bojanowski, P., Joulin, A.: {Unsupervised learning of visual features by contrasting cluster assignments}. In: NeurIPS (2020)

\bibitem{caron2021emerging}
Caron, M., Touvron, H., Misra, I., J{\'e}gou, H., Mairal, J., Bojanowski, P., Joulin, A.: {Emerging properties in self-supervised vision transformers}. In: CVPR (2021)

\bibitem{chen2023pixart}
Chen, J., Yu, J., Ge, C., Yao, L., Xie, E., Wu, Y., Wang, Z., Kwok, J., Luo, P., Lu, H., et~al.: {PixArt-$\alpha$: Fast Training of Diffusion Transformer for Photorealistic Text-to-Image Synthesis}. In: ICLR (2024)

\bibitem{choi2018stargan}
Choi, Y., Choi, M., Kim, M., Ha, J.W., Kim, S., Choo, J.: {StarGAN: Unified Generative Adversarial Networks for Multi-Domain Image-to-Image Translation}. In: CVPR (2018)

\bibitem{cocchi2023unveiling}
Cocchi, F., Baraldi, L., Poppi, S., Cornia, M., Baraldi, L., Cucchiara, R.: {Unveiling the Impact of Image Transformations on Deepfake Detection: An Experimental Analysis}. In: ICIAP (2023)

\bibitem{Corvi_2023_CVPR}
Corvi, R., Cozzolino, D., Poggi, G., Nagano, K., Verdoliva, L.: {Intriguing Properties of Synthetic Images: From Generative Adversarial Networks to Diffusion Models}. In: CVPR Workshops (2023)

\bibitem{corvi2023detection}
Corvi, R., Cozzolino, D., Zingarini, G., Poggi, G., Nagano, K., Verdoliva, L.: {On the detection of synthetic images generated by diffusion models}. In: ICASSP (2023)

\bibitem{cozzolino2018forensictransfer}
Cozzolino, D., Thies, J., R{\"o}ssler, A., Riess, C., Nie{\ss}ner, M., Verdoliva, L.: {ForensicTransfer: Weakly-supervised Domain Adaptation for Forgery Detection}. arXiv preprint arXiv:1812.02510  (2018)

\bibitem{5206848}
Deng, J., Dong, W., Socher, R., Li, L.J., Li, K., Fei-Fei, L.: {ImageNet: A large-scale hierarchical image database}. In: CVPR (2009)

\bibitem{dhariwal2021diffusion}
Dhariwal, P., Nichol, A.: {Diffusion Models Beat GANs on Image Synthesis}. In: NeurIPS (2021)

\bibitem{ding2021cogview}
Ding, M., Yang, Z., Hong, W., Zheng, W., Zhou, C., Yin, D., Lin, J., Zou, X., Shao, Z., Yang, H., et~al.: {CogView: Mastering Text-to-Image Generation via Transformers}. In: NeurIPS (2021)

\bibitem{dosovitskiy2020image}
Dosovitskiy, A., Beyer, L., Kolesnikov, A., Weissenborn, D., Zhai, X., Unterthiner, T., Dehghani, M., Minderer, M., Heigold, G., Gelly, S., et~al.: {An Image is Worth 16x16 Words: Transformers for Image Recognition at Scale}. In: ICLR (2021)

\bibitem{douze2024faiss}
Douze, M., Guzhva, A., Deng, C., Johnson, J., Szilvasy, G., Mazaré, P.E., Lomeli, M., Hosseini, L., Jégou, H.: {The Faiss Library}. arXiv preprint arXiv:2401.08281  (2024)

\bibitem{epstein2023online}
Epstein, D.C., Jain, I., Wang, O., Zhang, R.: {Online Detection of AI-Generated Images}. In: ICCV Workshops (2023)

\bibitem{esser2021taming}
Esser, P., Rombach, R., Ommer, B.: {Taming Transformers for High-Resolution Image Synthesis}. In: CVPR (2021)

\bibitem{frank2020leveraging}
Frank, J., Eisenhofer, T., Sch{\"o}nherr, L., Fischer, A., Kolossa, D., Holz, T.: {Leveraging frequency analysis for deep fake image recognition}. In: ICML (2020)

\bibitem{goodfellow2014generative}
Goodfellow, I., Pouget-Abadie, J., Mirza, M., Xu, B., Warde-Farley, D., Ozair, S., Courville, A., Bengio, Y.: {Generative adversarial nets}. In: NeurIPS (2014)

\bibitem{gragnaniello2021gan}
Gragnaniello, D., Cozzolino, D., Marra, F., Poggi, G., Verdoliva, L.: {Are GAN generated images easy to detect? A critical analysis of the state-of-the-art}. In: ICME (2021)

\bibitem{he2016identity}
He, K., Zhang, X., Ren, S., Sun, J.: {Identity Mappings in Deep Residual Networks}. In: ECCV (2016)

\bibitem{ho2020denoising}
Ho, J., Jain, A., Abbeel, P.: {Denoising diffusion probabilistic models}. In: NeurIPS (2020)

\bibitem{hong2023improving}
Hong, S., Lee, G., Jang, W., Kim, S.: Improving sample quality of diffusion models using self-attention guidance. In: ICCV (2023)

\bibitem{johnson2019billion}
Johnson, J., Douze, M., J{\'e}gou, H.: Billion-scale similarity search with {GPUs}. IEEE Trans. on Big Data  \textbf{7} (2019)

\bibitem{karras2017progressive}
Karras, T., Aila, T., Laine, S., Lehtinen, J.: {Progressive Growing of GANs for Improved Quality, Stability, and Variation}. In: ICLR (2018)

\bibitem{liao2022text}
Liao, W., Hu, K., Yang, M.Y., Rosenhahn, B.: {Text to image generation with semantic-spatial aware gan}. In: CVPR (2022)

\bibitem{loshchilov2017decoupled}
Loshchilov, I., Hutter, F.: Decoupled weight decay regularization. arXiv preprint arXiv:1711.05101  (2017)

\bibitem{DBLP:conf/iclr/LoshchilovH19}
Loshchilov, I., Hutter, F.: {Decoupled Weight Decay Regularization}. In: ICLR (2019)

\bibitem{van2008visualizing}
Van~der Maaten, L., Hinton, G.: Visualizing data using t-sne. Journal of machine learning research  \textbf{9}(11) (2008)

\bibitem{nichol2021glide}
Nichol, A., Dhariwal, P., Ramesh, A., Shyam, P., Mishkin, P., McGrew, B., Sutskever, I., Chen, M.: {GLIDE: Towards Photorealistic Image Generation and Editing with Text-Guided Diffusion Models}. In: ICML (2022)

\bibitem{ojha2023towards}
Ojha, U., Li, Y., Lee, Y.J.: {Towards Universal Fake Image Detectors That Generalize Across Generative Models}. In: CVPR (2023)

\bibitem{oord2018representation}
Oord, A.v.d., Li, Y., Vinyals, O.: {Representation learning with contrastive predictive coding}. arXiv preprint arXiv:1807.03748  (2018)

\bibitem{oquab2023dinov2}
Oquab, M., Darcet, T., Moutakanni, T., Vo, H.V., Szafraniec, M., Khalidov, V., Fernandez, P., Haziza, D., Massa, F., El-Nouby, A., Howes, R., Huang, P.Y., Xu, H., Sharma, V., Li, S.W., Galuba, W., Rabbat, M., Assran, M., Ballas, N., Synnaeve, G., Misra, I., Jegou, H., Mairal, J., Labatut, P., Joulin, A., Bojanowski, P.: {DINOv2: Learning Robust Visual Features without Supervision}. arXiv preprint arXiv:2304.07193  (2023)

\bibitem{patil2024amused}
Patil, S., Berman, W., Rombach, R., von Platen, P.: amused: An open muse reproduction. arXiv preprint arXiv:2401.01808  (2024)

\bibitem{scikit-learn}
Pedregosa, F., Varoquaux, G., Gramfort, A., Michel, V., Thirion, B., Grisel, O., Blondel, M., Prettenhofer, P., Weiss, R., Dubourg, V., Vanderplas, J., Passos, A., Cournapeau, D., Brucher, M., Perrot, M., Duchesnay, E.: {Scikit-learn: Machine Learning in {P}ython}. JMLR  \textbf{12} (2011)

\bibitem{podell2023sdxl}
Podell, D., English, Z., Lacey, K., Blattmann, A., Dockhorn, T., M{\"u}ller, J., Penna, J., Rombach, R.: {SDXL: Improving Latent Diffusion Models for High-Resolution Image Synthesis}. arXiv preprint arXiv:2307.01952  (2023)

\bibitem{radford2021learning}
Radford, A., Kim, J.W., Hallacy, C., Ramesh, A., Goh, G., Agarwal, S., Sastry, G., Askell, A., Mishkin, P., Clark, J., et~al.: {Learning transferable visual models from natural language supervision}. In: ICML (2021)

\bibitem{ramesh2022hierarchical}
Ramesh, A., Dhariwal, P., Nichol, A., Chu, C., Chen, M.: {Hierarchical Text-Conditional Image Generation with CLIP Latents}. arXiv preprint arXiv:2204.06125  (2022)

\bibitem{ramesh2021zero}
Ramesh, A., Pavlov, M., Goh, G., Gray, S., Voss, C., Radford, A., Chen, M., Sutskever, I.: {Zero-shot text-to-image generation}. In: ICML (2021)

\bibitem{razzhigaev2023kandinsky}
Razzhigaev, A., Shakhmatov, A., Maltseva, A., Arkhipkin, V., Pavlov, I., Ryabov, I., Kuts, A., Panchenko, A., Kuznetsov, A., Dimitrov, D.: {Kandinsky: an Improved Text-to-Image Synthesis with Image Prior and Latent Diffusion}. arXiv preprint arXiv:2310.03502  (2023)

\bibitem{rombach2022high}
Rombach, R., Blattmann, A., Lorenz, D., Esser, P., Ommer, B.: {High-resolution image synthesis with latent diffusion models}. In: CVPR (2022)

\bibitem{rossler2019faceforensics++}
Rossler, A., Cozzolino, D., Verdoliva, L., Riess, C., Thies, J., Nie{\ss}ner, M.: {Faceforensics++: Learning to detect manipulated facial images}. In: ICCV (2019)

\bibitem{saharia2022photorealistic}
Saharia, C., Chan, W., Saxena, S., Li, L., Whang, J., Denton, E.L., Ghasemipour, K., Gontijo~Lopes, R., Karagol~Ayan, B., Salimans, T., et~al.: {Photorealistic text-to-image diffusion models with deep language understanding}. In: NeurIPS (2022)

\bibitem{sauer2023adversarial}
Sauer, A., Lorenz, D., Blattmann, A., Rombach, R.: {Adversarial Diffusion Distillation}. arXiv preprint arXiv:2311.17042  (2023)

\bibitem{NIPS1999_8725fb77}
Sch\"{o}lkopf, B., Williamson, R.C., Smola, A., Shawe-Taylor, J., Platt, J.: Support vector method for novelty detection. In: NeurIPS (1999)

\bibitem{schuhmann2021laion}
Schuhmann, C., Vencu, R., Beaumont, R., Kaczmarczyk, R., Mullis, C., Katta, A., Coombes, T., Jitsev, J., Komatsuzaki, A.: {LAION-400M: Open Dataset of CLIP-Filtered 400 Million Image-Text Pairs}. In: NeurIPS Workshops (2021)

\bibitem{sha2022fake}
Sha, Z., Li, Z., Yu, N., Zhang, Y.: {DE-FAKE: Detection and Attribution of Fake Images Generated by Text-to-Image Generation Models}. In: ACM CCS (2023)

\bibitem{sohl2015deep}
Sohl-Dickstein, J., Weiss, E., Maheswaranathan, N., Ganguli, S.: {Deep Unsupervised Learning using Nonequilibrium Thermodynamics}. In: ICML (2015)

\bibitem{tao2023galip}
Tao, M., Bao, B.K., Tang, H., Xu, C.: {GALIP: Generative Adversarial CLIPs for Text-to-Image Synthesis}. In: CVPR (2023)

\bibitem{touvron2021training}
Touvron, H., Cord, M., Douze, M., Massa, F., Sablayrolles, A., J{\'e}gou, H.: {Training data-efficient image transformers \& distillation through attention}. In: ICML (2021)

\bibitem{ijcai2020p476}
Wang, R., Juefei-Xu, F., Ma, L., Xie, X., Huang, Y., Wang, J., Liu, Y.: {FakeSpotter: A Simple yet Robust Baseline for Spotting AI-Synthesized Fake Faces}. In: IJCAI (2020)

\bibitem{wang2020cnn}
Wang, S.Y., Wang, O., Zhang, R., Owens, A., Efros, A.A.: {CNN-Generated Images Are Surprisingly Easy to Spot... for Now}. In: CVPR (2020)

\bibitem{wang2023dire}
Wang, Z., Bao, J., Zhou, W., Wang, W., Hu, H., Chen, H., Li, H.: {DIRE for Diffusion-Generated Image Detection}. In: ICCV (2023)

\bibitem{wang-etal-2023-diffusiondb}
Wang, Z.J., Montoya, E., Munechika, D., Yang, H., Hoover, B., Chau, D.H.: {{D}iffusion{DB}: A Large-scale Prompt Gallery Dataset for Text-to-Image Generative Models}. In: ACL (2023)

\bibitem{yang2019exposing}
Yang, X., Li, Y., Lyu, S.: {Exposing Deep Fakes Using Inconsistent Head Poses}. In: ICASSP (2019)

\bibitem{yu2022scaling}
Yu, J., Xu, Y., Koh, J.Y., Luong, T., Baid, G., Wang, Z., Vasudevan, V., Ku, A., Yang, Y., Ayan, B.K., et~al.: Scaling autoregressive models for content-rich text-to-image generation. Trans. on Machine Learning Research  (2022)

\end{thebibliography}

\appendix
\section*{\vspace{0.2cm}Supplementary Material}
In the following, we provide additional implementation details and experimental results for the proposed model. Moreover, we also report a comprehensive analysis of the data generation process for the \dataset dataset.

\section{Additional Experimental Analyses\vspace{-0.1cm}}
\tinytit{Inference time analysis} 
As outlined in the main paper, one of the main benefits of training \ours from scratch is that this allows for more freedom in architectural choices. This ultimately results in a smaller and more efficient model compared to state-of-the-art detectors.

\begin{table}[b]
\vspace{-0.3cm}
  \caption{Parameter count and throughput of different deepfake detectors. Results are evaluated on a single RTX 6000 GPU equipped with 24 GB of VRAM, considering both batched and individual images. For each model, we take the largest possible batch size.}
  \label{tab:inference_time}
  \vspace{-0.2cm}
  \centering
  \setlength{\tabcolsep}{.4em}
  \resizebox{0.85\linewidth}{!}{
  \begin{tabular}{lc cc cc c cc}
    \toprule
    & & & & \multicolumn{2}{c}{\textbf{Single Sample}} & & \multicolumn{2}{c}{\textbf{Batched Samples}} \\
    & & & & \multicolumn{2}{c}{\textbf{Images/s} $\uparrow$} & & \multicolumn{2}{c}{\textbf{Images/s} $\uparrow$} \\
    \cmidrule{5-6} \cmidrule{8-9}
    \textbf{Model} & Params & Input size & & Backbone & All (NN) & & Backbone & All (NN)  \\
    \midrule
    Corvi~\etal~\cite{corvi2023detection} & 23 M & 224$^2$ & & 62.5 & - & & - & - \\
    Corvi~\etal~\cite{corvi2023detection} & 23 M & 512$^2$ & & 12.5 & - & & - & - \\
    Corvi~\etal~\cite{corvi2023detection} & 23 M & 768$^2$ & & 5.5 & - & & - & - \\
    Wang~\etal~(RN50)~\cite{wang2020cnn} & 26 M & 224$^2$ & & 196.0 & - & & 1023.8 & - \\
    \midrule
    DINOv2 ViT-B & 86 M & 518$^2$ & & 32.1 & 14.1 & & 35.2 & 15.8 \\
    Ojha~\etal~(CLIP ViT-L)~\cite{ojha2023towards} & 307 M & 224$^2$ & & 48.2 & 29.8 & & 68.0 & 59.8 \\
    CLIP ViT-B & 86 M & 224$^2$ & & 166.1 & 66.4 & & 289.8 & 234.0 \\
    \rowcolor{teagreen}
    \textbf{\ours~(ViT-T)} & 5 M & 224$^2$ & & \textbf{224.7} & \textbf{129.9} & & \textbf{2905.6} & \textbf{1729.1} \\
    \bottomrule
  \end{tabular}
}
\vspace{-0.2cm}
\end{table}

To showcase this, in Table~\ref{tab:inference_time} we compare \ours with the models proposed by Wang~\etal~\cite{wang2020cnn}, Corvi~\etal~\cite{corvi2023detection}, and Ojha~\etal~\cite{ojha2023towards} in terms of number of parameters, expected input size, and inference throughput (\ie~number of processed images per second). In the comparison, we also include CLIP DINOv2 with ViT-B as backbones. Specifically, we assess the throughput of the different architectures according to two settings, the former involving the evaluation of a single image at a time (``single sample'') and the latter considering the batching of multiple images to expedite the evaluation process (``batched samples''). Within these two setups, we evaluate performances both with and without the incorporation of a final nearest neighbor classifier~\cite{douze2024faiss}. Experiments are performed on a single RTX 6000 GPU equipped with 24 GB of VRAM. In the batched case, we identify the largest possible batch size (among powers of two) and compute the average number of images processed per second over 100 iterations.

As it can be observed, \ours is considerably faster than all the compared models. Notably, when testing the detector of Corvi~\etal~\cite{corvi2023detection} we do not crop or resize the image to a fixed size, following the original implementation. However, the model experiences substantial throughput degradation when subjected to high-resolution images. Specifically, the classification of images at a resolution of 768$^2$ results in a throughput of only 5.5 images per second, in contrast to a throughput of 62.5 images per second achieved when the images are resized to 256$^2$. Additionally, the evaluation of images in their unaltered resolutions introduces challenges, particularly in batch-processing contexts, due to the potential for varying aspect ratios among images. This variability necessitates limiting the batch size to 1, further impacting the throughput efficiency of the model. 

Conversely, \ours can process up to 129.9 images per second when feeding single images to the model, which is 2.0 and 4.4 times faster than, respectively, the CLIP ViT-B and CLIP ViT-L models employed in~\cite{ojha2023towards}. When batching the input images, moreover, \ours reaches a throughput of 1729.1 images per second, which amounts to a speed-up of 7.4 and 28.9 times in comparison to the aforementioned models.

\begin{table*}[t]
  \caption{Contribution of each loss component within our proposed methodologies on \dataset extended test set. We report accuracy scores over real images and fake ones from each of the 12 generators included in the test set. Additionally, we include the average accuracy across all fake images (Avg) and those generated by diffusion models not seen during training (Avg$_\text{u}$). The $\dagger$ marker indicates methods trained on other datasets and tested on the \dataset extended test set.}
  \label{tab:ablation_extended_test_set}
  \vspace{-0.2cm}
  \centering
  \setlength{\tabcolsep}{.18em}
  \resizebox{\linewidth}{!}{
  \begin{tabular}{lc cc cccccccccccc cc}
    \toprule
    \textbf{Model} & & Real & & DF-IF & SD-1.4 & SD-2.1 & SD-XL & SD-XL-T & unCLIP & SAG & aMUSEd & K-2.1 & K-2.2 & K-3 & PA-$\alpha$ & \textbf{Avg} & \textbf{Avg$_\text{u}$} \\
    \midrule
    w/ $\mathcal{L}_\text{global}$ only (real $\rightarrow$ fake) & &  72.4 & & \textbf{85.7} & 86.6 & 88.5 & 76.1 & 65.2 & \textbf{97.7} & 93.9 & 59.8 & 57.8 & 61.9 & 49.4 & 59.2 & 73.5 & 68.1 \\
    w/ $\mathcal{L}_\text{global}$ only (pre-trained) & & 90.4 & & 57.8 & 62.1 & 65.4 & 61.3 & 63.1 & 84.7 & 84.1 & 55.8 & 41.0 & 60.1 & 40.8 & 50.8 & 60.6 & 60.0 \\
    \rowcolor{teagreen}
    \textbf{\ours (Linear)} & & \textbf{97.3} & & 85.2 & \textbf{98.8 }& \textbf{98.7} & \textbf{92.9}& \textbf{86.4} & \textbf{\textbf{95.0}} & \textbf{99.2} & \textbf{72.8} & \textbf{70.5} & \textbf{82.7} & \textbf{67.1} & \textbf{86.5} & \textbf{86.3 }& \textbf{82.5} \\
    \midrule
    w/ $\mathcal{L}_\text{global}$ only (real $\rightarrow$ fake) & & 71.9 & & 72.3 & 69.2 & 70.0 & 67.2 & 52.5 & 85.2 & 77.2 & 51.6 & 50.9 & 52.2 & 47.3 & 47.8 & 61.9 & 58.1 \\
    w/ $\mathcal{L}_\text{global}$ only (pre-trained) & & 82.6 & & 66.5 & 63.7 & 67.2 & 69.6 & 70.9 & 83.8 & 84.4 & 66.3 & 49.5 & 66.1 & 47.7 & 57.2 & 66.1 & 65.7 \\
    \rowcolor{teagreen}
    \textbf{\ours (NN)} & & \textbf{95.3} & & \textbf{90.2} &\textbf{ 99.3} & \textbf{99.2} & \textbf{95.1} & \textbf{91.2} & \textbf{97.2} & \textbf{99.5} & \textbf{81.4} & \textbf{78.5} & \textbf{87.5} & \textbf{75.8} & \textbf{91.1} & \textbf{90.5} &\textbf{ 87.8} \\
    \midrule
    w/ $\mathcal{L}_\text{global}$ only (real $\rightarrow$ fake) & & 88.3 & & 71.2 & 63.8 & 66.6 & 52.2 & 46.5 & 91.8 & 76.9 & 40.7 & 37.6 & 35.9 & 27.9 & 36.2 & 53.9 & 49.2 \\
    w/ $\mathcal{L}_\text{global}$ only (pre-trained) & & 85.9 & & 53.5 & 57.4 & 59.2 & 78.1 & 63.0 & 50.1 & 73.6 & 49.3 & 40.4 & 63.6 & 51.3 & 41.2 & 56.7 & 54.1 \\
    \rowcolor{teagreen}
    \textbf{\ours (SVM)} & & \textbf{91.9} & & \textbf{96.9} & \textbf{91.0} & \textbf{94.8} & \textbf{98.6} & \textbf{97.0} & \textbf{93.0} & \textbf{95.1} & \textbf{94.4} & \textbf{92.2} & \textbf{97.5} & \textbf{92.2} & \textbf{95.2} & \textbf{94.8} & \textbf{94.7} \\

    \bottomrule
  \end{tabular}
}
\vspace{-0.3cm}
\end{table*}

\tit{Additional ablation studies}
In Table~\ref{tab:ablation_extended_test_set}, we present the results of the ablation study conducted on an expanded test set derived from the \dataset, focusing on the individual components of our loss function. Despite modifications to the generators, the observed results exhibit patterns similar to those reported in Table~\ref{tab:ablation}, demonstrating consistent behavior. For instance, when \ours is augmented with a linear classifier, it exhibits an average accuracy enhancement on unseen generators of 22.5\% and 14.4\% over the baseline pre-trained model and when solely employing $\mathcal{L}_\text{global}$ (real $\rightarrow$ fake), respectively. Remarkably, these enhancements increase to 40.6\% and 45.5\% in scenarios leveraging one-class SVM classifiers, compared to the aforementioned models.

\begin{table*}[t]
  \caption{
  Performance evaluation of methods re-trained on \dataset and tested on data considering settings with and without image transformations. 
  “Overall” accuracy averages predictions for both fake and real
  images. “Fake” accuracy denotes the mean accuracy for generated data. 
  }
  \label{tab:retrained_all_generators}
  \vspace{-0.2cm}
  \centering
  \setlength{\tabcolsep}{.32em}
  \resizebox{0.95\linewidth}{!}{
  \begin{tabular}{lc ccc cc ccc c cccc}
    \toprule
    & & \multicolumn{3}{c}{\textbf{w/o Transforms}} & & & \multicolumn{8}{c}{\textbf{w/ Transforms}} \\
    \cmidrule(lr){3-5} \cmidrule(lr){8-15}
    \textbf{Model} & & Overall & Real & Fake & & & Overall & Real & Fake & & DF-IF & SD-1.4 & SD-2.1 & SD-XL \\
    \midrule
    Wang~\etal~\cite{wang2020cnn} & & \textbf{99.4} & 98.6 & 99.6 & & & 83.6 & 99.4 & 79.7 & & 74.3 & 81.4 & 80.5 & 82.5 \\   
    Gragnaniello/Corvi~\etal~\cite{gragnaniello2021gan,corvi2023detection} & & \underline{93.3} & 96.7 & 92.4 & & & 75.0 & 98.9 & 69.0 & & 66.0 & 69.4 & 67.5 & 72.9 \\
    Ojha~\etal~\cite{ojha2023towards} & & 89.9 & 81.2 & 92.2 & & & 81.5 & 91.8 & 78.9 & & 71.5 & 78.5 & 78.3 & 87.3 \\
    \midrule
    \rowcolor{teagreen}
    \textbf{\ours (Linear)} & & 98.0 & 94.0 & 99.0 & & & 95.7 & 95.6 & \underline{95.8} & & 94.8 & 95.8 & 94.9 & 97.5 \\
    \rowcolor{teagreen}
    \textbf{\ours (NN)} & & 97.3 & 89.3 & 99.3 & & & \textbf{95.8} & 90.5 & 97.1 & & 96.6 & 97.3 & 96.6 & 98.1 \\
    \rowcolor{teagreen}
    \textbf{\ours (SVM)} & & 91.2 & 74.4 & 95.4 & & & 92.5 & 81.0 & 95.4 & & 96.6 & 91.7 & 94.4 & 99.0 \\
    \bottomrule
  \end{tabular}
}
\vspace{-0.1cm}
\end{table*}

\begin{table*}[t]
  \caption{Performance evaluation of methods re-trained on \dataset and tested on \dataset extended test set. We report accuracy scores over real images and fake ones from each of the 12 generators included in the test set. Additionally, we include the average accuracy across all fake images (Avg) and those generated by diffusion models not seen during training (Avg$_\text{u}$).}
  \label{tab:retrained_extended_test_set}
  \vspace{-0.2cm}
  \centering
  \setlength{\tabcolsep}{.18em}
  \resizebox{\linewidth}{!}{
  \begin{tabular}{lc cc cccccccccccc cc}
    \toprule
    \textbf{Model} & & Real & & DF-IF & SD-1.4 & SD-2.1 & SD-XL & SD-XL-T & unCLIP & SAG & aMUSEd & K-2.1 & K-2.2 & K-3 & PA-$\alpha$ & \textbf{Avg} & \textbf{Avg$_\text{u}$} \\
    \midrule
    Wang~\etal~\cite{wang2020cnn} & &  97.0 & & 72.8 & 100.0 & 100.0 & 98.4 & 87.9 & 97.0 & 99.9 & 70.0 & 85.6 & 98.0 & 89.4 & 97.6 & \underline{91.4} & \underline{90.7} \\
    
    Gragnaniello/Corvi~\etal~\cite{gragnaniello2021gan,corvi2023detection} & & 81.5 & & 74.5 & 98 & 98 & 97.1 & 91.1 & 92.8 & 96.7 & 85.3 & 77.1 & 92.4 & 84.7 & 94.6 & 90.2 & 89.3 \\
    
    Ojha~\etal~\cite{ojha2023towards} & & 70.1 & & 88.2 & 86.7 & 87.1 & 86.0 & 96.1 & 80.6 & 91.7 & 95.4 & 88.5 & 92.8 & 85.9 & 94.4 & 89.5 & 90.7 \\

    \midrule
    \rowcolor{teagreen}
    \textbf{\ours (Linear)} & & 97.3 & & 85.2 & 98.8 & 98.7 & 92.9 & 86.4 & 95.0 & 99.2 & 72.8 & 70.5 & 82.8 & 67.1 & 86.5 & 86.3 & 82.5 \\
    \rowcolor{teagreen}
    \textbf{\ours (NN)} & & 95.3 & & 90.2 & 99.3 & 99.2 & 95.1 & 91.2 & 97.2 & 99.5 & 81.4 & 78.5 & 87.5 & 75.8 & 91.1 & 90.5 & 87.8 \\
     \rowcolor{teagreen}
    \textbf{\ours (SVM)} & & 91.9 & & 96.9 & 91.0 & 94.8 & 98.6 & 97.0 & 93.0 & 95.1 & 94.4 & 92.2 & 97.5 & 92.2 & 96.2 & \textbf{94.8} & \textbf{94.7} \\
    \bottomrule
  \end{tabular}
}
\vspace{-0.3cm}
\end{table*}
\begin{table}[t]
  \caption{Accuracy results of methods re-trained on \dataset and tested on external unseen generators employing data from a mixture of diffusion and autoregressive models. Following~\cite{ojha2023towards}, we average the accuracy scores between real and fake images.}
  \label{tab:retrained_datasets_diffusion}
    \vspace{-0.2cm}
  \centering
  \setlength{\tabcolsep}{.25em}
  \resizebox{\linewidth}{!}{
  \begin{tabular}{lc c c ccc c ccc c ccc c c c c}
    \toprule
    & & & & \multicolumn{3}{c}{LDM} & & \multicolumn{3}{c}{GLIDE} & & \multicolumn{3}{c}{DALL-E} \\
    \cmidrule{5-7} \cmidrule{9-11} \cmidrule{13-15}
   \textbf{Model} & & Guided & & 200 & 200 (CFG) & 100 & & 100 (27) & 50 (27) & 100 (10) & & v1 & v2 & v3 & & Midjourney & & \textbf{Avg} \\
    \midrule
    Wang~\etal~\cite{wang2020cnn} & & 51.7 & & 99.9 & 99.6 & 99.7 & & 62.3 & 65.4 & 64.4 & & 52.1 & 66.7 & 86.6 & & 96.5 & & 76.8 \\
    Ojha~\etal~\cite{ojha2023towards} & & 56.8 & & 76.6 & 74.8 & 75.2 & & 80.2 & 80.4 & 82.1 & & 84.4 & 93.3 & 88.5 & & 76.4 & & 79.0 \\
    Gragnaniello/Corvi~\etal~\cite{gragnaniello2021gan,corvi2023detection} & & 54.6 & & 98.8 & 99.6 & 99.2 & & 67.4 & 71.2 & 69.1 & & 57.5 & 80.6 & 73.8 & & 93.0 & & 78.6 \\
    
    \midrule
    \rowcolor{teagreen}
    \textbf{\ours (Linear)} & & 53.5 & & 92.5 & 95.6 & 91.9 & & 71.7 & 75.4 & 72.9 & & 63.1 & 71.4 & 86.7 & & 84.0 & & 78.0 \\
    \rowcolor{teagreen}
    \textbf{\ours (NN)} & & 53.5 & & 92.7 & 96.1 & 92.5 & & 73.8 & 76.9 & 74.0 & & 67.0 & 74.3 & 88.6 & & 86.8 & & \underline{79.6} \\
    \rowcolor{teagreen}
    \textbf{\ours (SVM)} & & 54.6 & & 91.0 & 90.4 & 90.9 & & 77.2 & 78.8 & 77.6 & & 76.1 & 80.2 & 91.0 & & 89.7 & & \textbf{81.6}\\
    \bottomrule
  \end{tabular}
}
\vspace{-0.1cm}
\end{table}

\tit{Comparison with other detectors trained on \dataset}
To expand the analysis on the \dataset dataset and increase the fairness of the evaluation setting, we train the some of the competitors considered in the main paper~\cite{corvi2023detection,gragnaniello2021gan,ojha2023towards,wang2020cnn} on the \dataset training set. During the training phase, we employ the original source codes when available or follow the training settings specified in the original methodologies. Due to the similarities between Corvi~\etal~\cite{corvi2023detection} and Gragnaniello~\etal~\cite{gragnaniello2021gan}, we consider only one model for both. 

We report accuracy results on three distinct test sets shown in the main paper to assess robustness to image transformations (\ie~Table~\ref{tab:retrained_all_generators}) and generalization to unseen generators (\ie~Tables~\ref{tab:retrained_extended_test_set} and~\ref{tab:retrained_datasets_diffusion}).
Noteworthy, in Table~\ref{tab:retrained_all_generators} all methods show low robustness on realistic settings like transformed images (w/ Transforms).
For instance, all \ours classifiers outperform the best-performing competitor method~\cite{wang2020cnn} by a margin of 10\% on ``Overall'' accuracy. Regarding instead the generalization to unseen generators, in Table~\ref{tab:retrained_extended_test_set} \ours SVM reaches higher average
accuracy results on both seen and unseen generators. Specifically, \ours gains 3.4\%, 4.6\%, and 5.3\% with respect to Wang~\etal~\cite{wang2020cnn}, Gragnaniello~\etal~\cite{gragnaniello2021gan}, and Ojha~\etal~\cite{ojha2023towards}. Similarly, in Table~\ref{tab:retrained_datasets_diffusion} while \ours Linear underperforms by 1.0\% on average accuracy compared to best competitor (\ie~Ojha~\etal), \ours NN and SVM obtain superior performance with respect to all the competitors by a margin of 0.6\% and 2.6\%.


\begin{table}[t]
    \caption{AUC-ROC results, averaged across the generators of each setting considered in the main paper (\ie~\dataset standard test set, \dataset extended test set, and additional external unseen generators).}
    \vspace{-0.2cm}
  \centering
  \setlength{\tabcolsep}{.26em}
  \resizebox{0.70\linewidth}{!}{
  \begin{tabular}{lc cc cc cc c}
    \toprule
    & & \multicolumn{3}{c}{\dataset} & & \dataset extended & & External generators \\
    \cmidrule{3-5} \cmidrule{7-7} \cmidrule{9-9}
    & & w/o T & & w / T & & Avg & & Avg \\
    \midrule
    Wang~\etal~(0.1)~\cite{wang2020cnn} & & 52.4 & & 52.3 & & 48.9 & & 72.1 \\
    Wang~\etal~(0.5)~\cite{wang2020cnn} & & 50.0 & & 50.0 & & 52.9 & & 65.7 \\
    Gragnaniello~\etal~\cite{gragnaniello2021gan} & & 68.8 & & 73.4 & & 68.4 & & 86.4 \\
    Corvi~\etal~\cite{corvi2023detection} & & 91.8 & & 87.1 & & 96.3 & & 90.7 \\
    Ojha~\etal~\cite{ojha2023towards} & & 64.4 & & 64.7 & & 69.6 & & 88.5 \\
    \midrule
    DINOv2 ViT-B (Linear) & & 92.4 & & 91.0 & & 80.6& & 86.5 \\
    CLIP ViT-L (Linear) & & \textbf{96.9} & & \underline{95.7} & & 85.0 & & \underline{91.3} \\
    \midrule
    \rowcolor{teagreen}
    \textbf{\ours (Linear)} & & 93.9 & & 93.7 & & \textbf{98.0} & & \textbf{92.2} \\
    \rowcolor{teagreen}
    \textbf{\ours (SVM)} & & \underline{95.9} & & \textbf{97.0} & & \underline{97.0} & & 90.5 \\
    \bottomrule
  \end{tabular}
}
\vspace{-0.3cm}

\label{tab:auc}
\end{table}

\tit{Detection performance on AUC-ROC metric} 
In addition to the standard evaluations in terms of accuracy, we evaluate our models using the AUC metric, typically used to assess unbalanced data distributions. However, calculating AUC for NN and one-class SVM classifiers is not possible as they do not provide a decision probability. We circumvent this for the one-class SVM by computing an outlier score: we calculate the maximum distance from the decision boundary in the training set and subtract it from the distance of the point being evaluated.
In Table~\ref{tab:auc}, we report the averaged AUC of \ours and competitors on all the evaluation settings (cf. Table~\ref{tab:all_generators}, Table~\ref{tab:extended_test_set}, and Table~\ref{tab:datasets_diffusion} of the main paper). Overall, \ours achieves superior performance in almost all settings. For instance, \ours Linear obtains a gain of 1.7\% and 1.5\% over one of the best competitors~\cite{corvi2023detection}. Further, \ours SVM obtains the second best performance in both Table~\ref{tab:extended_test_set} and Table~\ref{tab:all_generators}. It is also worth noting that, except for the ``All'' columns of Table~\ref{tab:all_generators}, all our evaluations are reported in terms of separate accuracy on real and fake images (Table~\ref{tab:extended_test_set}) or averaged accuracy on the same number of real and fake data (Table~\ref{tab:datasets_diffusion}), thus not having imbalanced settings.

\begin{table*}[t]
  \caption{Accuracy results on external unseen generators employing data from different generative adversarial networks. Following~\cite{ojha2023towards}, we average the accuracy scores between real and fake images.
  }
  \label{tab:gans}
  \vspace{-0.2cm}
  \centering
  \setlength{\tabcolsep}{.4em}
  \resizebox{0.85\linewidth}{!}{
  \begin{tabular}{lc cccccc cc}
    \toprule
    \textbf{Model} & & ProGAN & CycleGAN &  BigGAN & StyleGAN & GauGAN & StarGAN & & \textbf{Avg} \\
    \midrule
    Corvi~\etal~\cite{corvi2023detection} & & 51.2 & 46.3 & 51.9 & 59.8 & 50.6 & 45.8 & & 50.9 \\
    Wang~\etal~(DIRE)~\cite{wang2023dire} & & 51.5 & 50.5 & 50.9 & 50.7 & 51.4 & 49.8 & & 50.8 \\
    \rowcolor{teagreen}
    \textbf{\ours (Linear)} & & \textbf{80.4} & \textbf{64.7} & \textbf{60.3} & \textbf{61.0} & \textbf{61.4} & \textbf{60.0} & & \textbf{64.5} \\
    \bottomrule
  \end{tabular}
}
\vspace{-0.1cm}
\end{table*}

\tit{Generalization analysis on GAN-generated images} To evaluate the performance on GAN-generated images, we consider the datasets provided by~\cite{ojha2023towards}, in this case limiting the analysis on fake images from GANs. Following the same setting of Table~\ref{tab:datasets_diffusion}, we report the average accuracy over both real and fake data. Results are reported in Table~\ref{tab:gans}, comparing our model with linear classifier with the two diffusion-based approaches considered in the main paper~\cite{corvi2023detection,wang2023dire}. In this case, the linear is fitted on 9,600 records of real and fake images respectively from ImageNet and ProGAN. Although the results achieved by our model are very distant from the ones obtained on images generated by diffusion models, a similar disparity is observed when comparing the accuracy scores of competitors, which struggle to accurately classify GAN-generated images. Nonetheless, it is essential to emphasize that our study primarily focuses on diffusion models, which currently represent the gold standard in generative AI. These models yield notably more realistic images compared to GANs, thus underscoring the significance of our research within this domain.

\begin{figure*}[t]
    \centering
    \begin{tabular}{c}
    \includegraphics[width=0.95\linewidth]{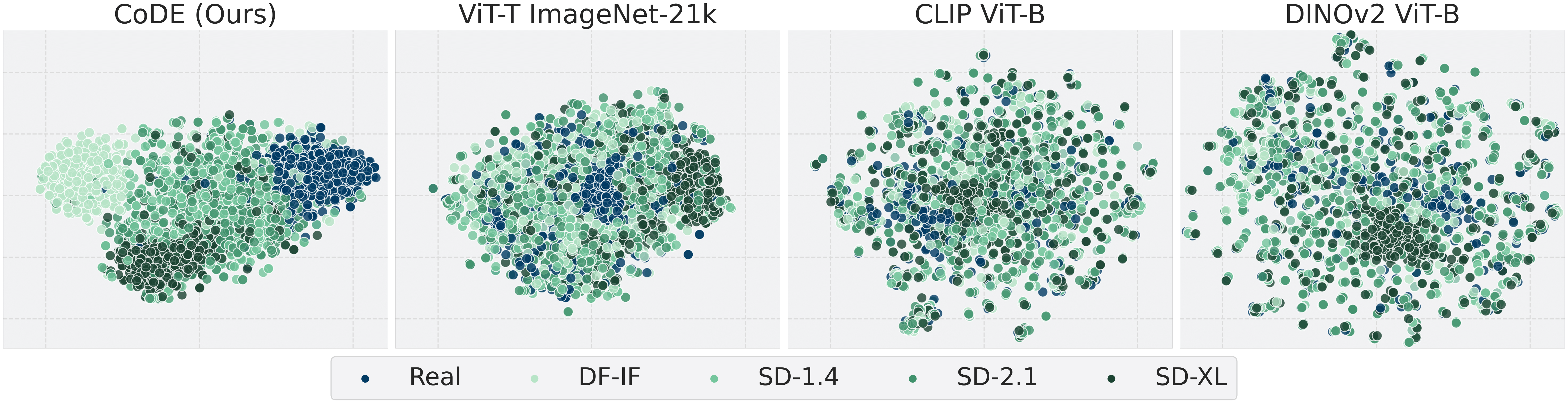} \\
    \addlinespace[0.15cm]
     \includegraphics[width=0.95\linewidth]{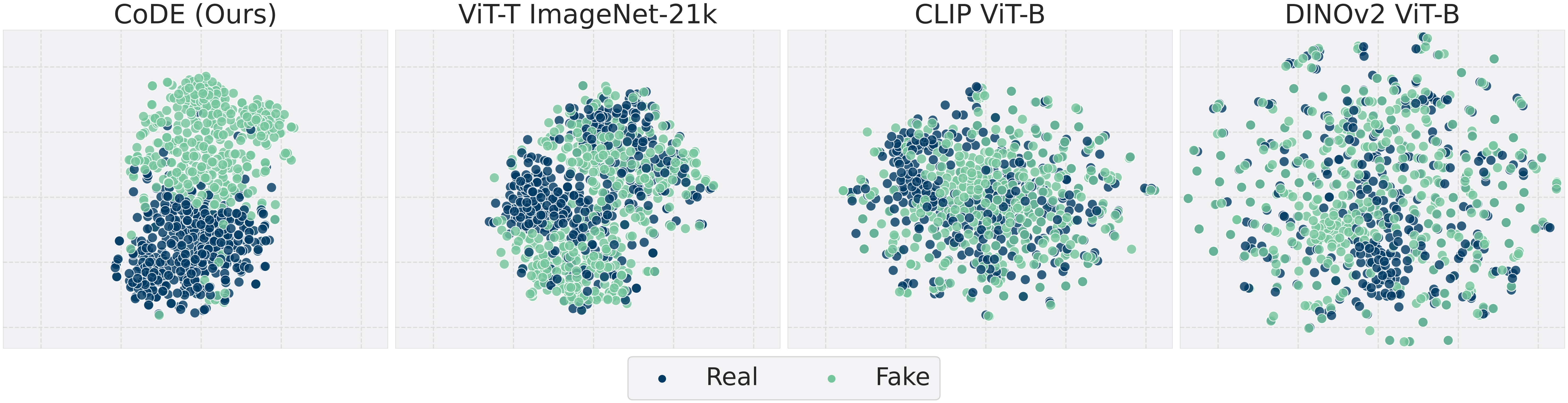}
    \end{tabular}
    \vspace{-.2cm}
    \caption{t-SNE visualizations of images without transformations, according to different backbones.}
    \label{fig:t-sne_suppl}
    \vspace{-.3cm}
\end{figure*}

\tit{Embedding space evaluation} 
In Fig.~\ref{fig:t-sne_suppl}, we report a t-SNE~\cite{van2008visualizing} visualization obtained with images from the test set of \dataset, encoded with different backbones. The provided plots facilitate the analysis of the embedding spaces, enabling the visualization of the image embeddings from various generators (first row) or the aggregation of distinct generators for a comprehensive real-fake comparison (second row). Notably, it can be observed how \ours tends to separate real and fake images in distinct regions of the embedding space. In contrast, pre-trained models such as ViT-T, CLIP ViT-B, and DINOv2 ViT-B do not separate between real and fake images, thus outlining that most of the information contained in their feature vectors is not useful for deepfake classification.

\begin{figure*}[t]
    \centering
     \begin{tabular}{c}
    \includegraphics[width=0.95\linewidth]{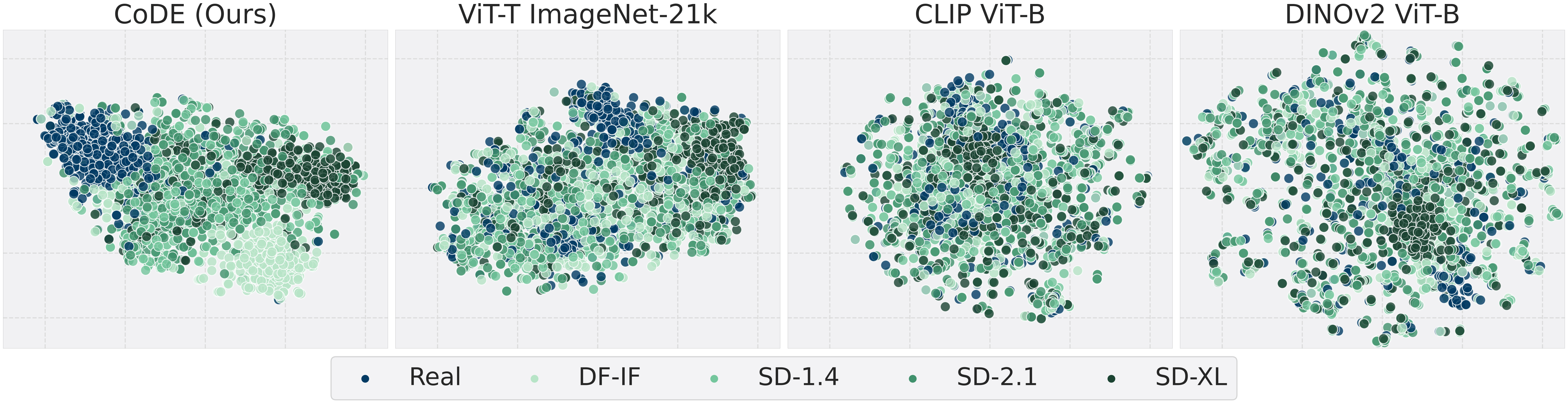}\\
    \addlinespace[0.15cm]
    \includegraphics[width=0.95\linewidth]{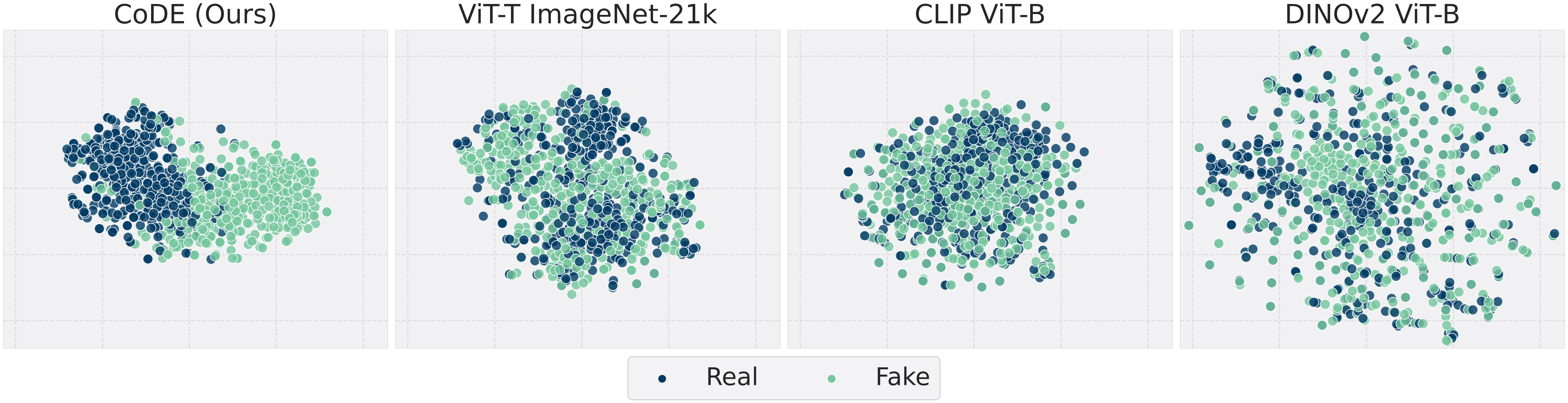}
   \end{tabular}
    \vspace{-.2cm}
    \caption{t-SNE visualizations of transformed images, according to different backbones.}
    \label{fig:t-sne_suppl_augm}
    \vspace{-.3cm}
\end{figure*}

Similar observations can be made by observing Fig.~\ref{fig:t-sne_suppl_augm}, where we also process input images with the transform operator $\mathcal{T}(\cdot)$. Notably, \ours maintains a clustered structure even when tasked with transformed images, thus exhibiting high invariance to realistic image transforms. In contrast, competitors exhibit representations characterized by a lack of spatial coherence. Overall, these findings confirm the appropriateness of training deepfake detection networks from scratch and with contrastive objectives that promote invariance to transformations and global-local mappings.

\section{Additional Implementation Details\vspace{-0.1cm}}
\tinytit{Images from commercial generative tools}
In Table~\ref{tab:datasets_diffusion} of the main, we present the performance metrics of \ours, alongside various other deepfake detection algorithms, when applied to images generated by state-of-the-art commercial generative models such as DALL-E 2~\cite{ramesh2022hierarchical}\footnote{\scriptsize\url{https://huggingface.co/datasets/SDbiaseval/jobs-dalle-2}}, DALL-E 3~\cite{betker2023improving}\footnote{\scriptsize\url{https://huggingface.co/datasets/OpenDatasets/dalle-3-dataset}}, and Midjourney V5\footnote{\scriptsize\url{https://huggingface.co/datasets/wanng/midjourney-v5-202304-clean}}. For the evaluation of each generative tool, a dataset comprising 1,000 images has been assembled utilizing publicly available collections. 

\tit{Backbone training} 
During \ours training, we employ a linear learning rate warmup strategy which starts from a learning rate of 1$e^{-6}$ and ends with a learning rate of 2$e^{-3}$ after five epochs. Then, a cosine learning rate schedule is employed.  
For optimization, we use the AdamW~\cite{loshchilov2017decoupled} optimizer with $\epsilon$ set to 1$e^{-8}$ and $\beta$ to $(0.9, 0.99)$. 
When training on \dataset, we employ a mini-batch size of 1,024 distributed across 4 GPUs, which amounts to processing 1,770 batches in each epoch. Moreover, we apply early stopping with patience set to 6, resulting in approximately 30 epochs of training. All the ViT backbones considered in this study share the same input size of $224^2$ and patch dimensionality of $16^2$, with the exception of CLIP ViT-L~\cite{radford2021learning} and DINOv2 ViT-B~\cite{oquab2023dinov2}. The latter two models employ $14^2$ patches, and in the case of the DINOv2 ViT-B model, an input size of $518^2$.

\tit{Classifier training} To evaluate the pre-trained backbones and \ours, we employ logistic regression~\cite{scikit-learn}, nearest neighbor classifier~\cite{johnson2019billion}, and one-class SVM~\cite{NIPS1999_8725fb77}. To better weigh the relevance of both real and fake data, we perform logistic regression with a balanced loss, training for a maximum of 1,500 iterations with an $\ell_2$ penalty and an inverse coefficient of regularization of $0.316$. 

\begin{figure*}[t]
\centering
\includegraphics[width=\linewidth]{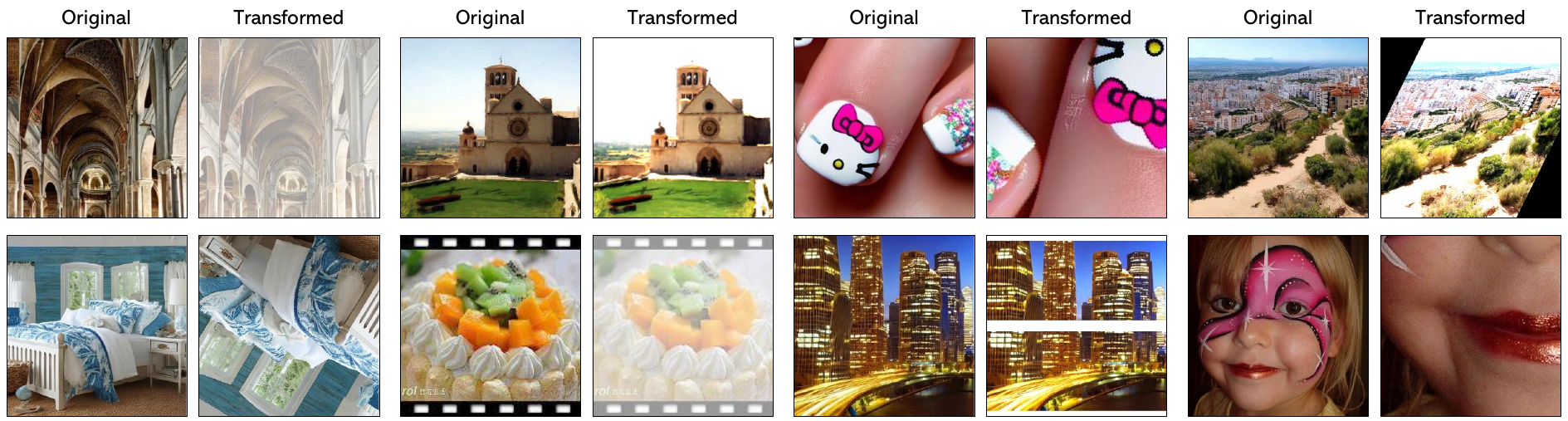}
\vspace{-.4cm}
\caption{Qualitative results of image transformations obtained with the transform operator $\mathcal{T}(\cdot)$.}
\label{fig:dataset_transf_supp}
\vspace{-.3cm}
\end{figure*}

\tit{Transformation protocol} 
As mentioned in Section~\ref{sec:method} of the main paper, the $\mathcal{T}(\cdot)$ operator allows us to ensure robustness to transformations by randomly sampling and applying multiple image transformations. In particular, $\mathcal{T}(\cdot)$ samples a random number of transformations (between zero and two), and all the selected transformations are applied with a randomized strength. To achieve this objective, we set a strength range for each augmentation, bounded by minimum and maximum values. The aim is to uphold visual quality in both scenarios, 
\begin{wraptable}{r}{0.48\textwidth}
\vspace{-0.8cm}
  \caption{Parameter ranges of the transformations employed in the $\mathcal{T}(\cdot)$ operator.}
  \label{tab:transformation}
  \vspace{0.1cm}
  \centering
  \setlength{\tabcolsep}{0.4em}
  \resizebox{0.98\linewidth}{!}{
  \begin{tabular}{lc cc c}
    \toprule
    & & \multicolumn{2}{c}{\textbf{Range}} \\
    \cmidrule{3-4}
    \textbf{Transformation} & & Min & Max  \\
    \midrule
    Contrast & & 0.5 & 1.5  \\
    Saturation & & 0.5 & 1.5  \\
    Encoding Quality Transform & & 40 & 100  \\
    Opacity & & 0.2 & 1.0  \\
    Overlay Stripes & & 0.05 & 0.35  \\
    Pad & & 0.01 & 0.25  \\
    Resize & & 64 & 512  \\
    Scale & & 0.5 & 1.5  \\
    Sharpen & & 1.2 & 2.0  \\
    Skew & & 1.0 & 1.0  \\
    Random Blur & & 0.1 & 2.0  \\
    Random Brightness & & 0.5 & 2.0  \\
    Random Aspect Ratio & & 0.75 & 2.0  \\
    Random Pixelization & & 0.3 & 1  \\
    Random Rotation & & 90 & 270  \\
    \midrule
    Grayscale & & - & -  \\
    Horizontal Flip & & - & -  \\
    \bottomrule
  \end{tabular}
}
  \vspace{-.5cm}
\end{wraptable}
ensuring the preservation of visual consistency and usability. The range of each transformation type is linearly partitioned into five equally spaced segments, yielding five distinct image augmentations for each transform, each exhibiting varying degrees of strength. When a transformation is sampled, also one of the five strength values is randomly selected and applied. The complete experimental configuration for the transform operator $\mathcal{T}(\cdot)$ is reported in Table~\ref{tab:transformation}. 

In our approach, each input image undergoes random cropping to $224^2$ during training, while center cropping is employed at inference time $224^2$. This step is applied across various image types, including real, fake, local, and global crop variations. Subsequently, the images undergo normalization using mean and standard deviation values computed from the ImageNet dataset. To qualitatively visualize the transformation process, we present in Fig.~\ref{fig:dataset_transf_supp} sample original and corresponding transformed images of the \dataset test set. 

\section{Additional Details on the \dataset Dataset\vspace{-0.1cm}} 
As outlined in Section~\ref{sec:dataset} of the main paper, the standard training and test splits of our \dataset dataset comprise images from four state-of-the-art generators. Additionally, we also collect an extended test set with images generated by 12 different diffusion-based models, including the four contained in the standard training and test sets. The specific models and model names employed from the \texttt{diffusers} library\footnote{\scriptsize\url{https://huggingface.co/docs/diffusers}} are reported in Table~\ref{tab:names}. 

\tit{Aspect ratios and sizes}
The aspect ratios and sizes of the images generated with the three Stable Diffusion models are subject to probabilistic sampling, with a 0.5 probability assigned to $512^2$ and a 0.25 probability assigned to $640\times480$ and $640\times360$.
When generating images with the DeepFloyd IF model, instead, we only perform the first two steps of its pipeline, thus generating an image with a size of $256^2$. Nevertheless, future works might still apply the upscaling model\footnote{\scriptsize\url{https://huggingface.co/stabilityai/stable-diffusion-x4-upscaler}} and obtain images with a size of $1024^2$ starting from samples contained in \dataset.

\begin{table}[t]
  \centering
  \caption{Model names from the \texttt{diffusers} library employed during the construction of the \dataset dataset.}
  \label{tab:names}
  \vspace{-0.2cm}
  \setlength{\tabcolsep}{.5em}
  \resizebox{0.8\linewidth}{!}{
  \begin{tabular}{lc l}
    \toprule
    \textbf{Model} & & \textbf{Model Name} \\
    \midrule
    DeepFloyd IF & & \footnotesize{\texttt{\href{https://huggingface.co/DeepFloyd/IF-II-L-v1.0}{DeepFloyd/IF-II-L-v1.0}}} \\
    Stable Diffusion 1.4 & & \footnotesize{\texttt{\href{https://huggingface.co/CompVis/stable-diffusion-v1-4}{CompVis/stable-diffusion-v1-4}}} \\
    Stable Diffusion 2.1 & & \footnotesize{\texttt{\href{https://huggingface.co/stabilityai/stable-diffusion-2-1-base}{stabilityai/stable-diffusion-2-1-base}}} \\
    Stable Diffusion XL & & \footnotesize{\texttt{\href{https://huggingface.co/stabilityai/stable-diffusion-xl-base-1.0}{stabilityai/stable-diffusion-xl-base-1.0}}} \\
    \midrule
    Stable Diffusion XL Turbo & & \footnotesize{\texttt{\href{https://huggingface.co/stabilityai/sdxl-turbo}{stabilityai/sdxl-turbo}}} \\
    Stable Diffusion unCLIP & & \footnotesize{\texttt{\href{https://huggingface.co/stabilityai/stable-diffusion-2-1-unclip-small}{stabilityai/stable-diffusion-2-1-unclip-small}}} \\
    Self-Attention Guidance & & \footnotesize{\texttt{\href{https://huggingface.co/docs/diffusers/api/pipelines/self_attention_guidance}{pipelines/selfattentionguidance}}} \\
    aMUSEd & & \footnotesize{\texttt{\href{https://huggingface.co/amused/amused-512}{amused/amused-512}}} \\
    Kandinsky 2.1 & & \footnotesize{\texttt{\href{https://huggingface.co/kandinsky-community/kandinsky-2-1-prior}{kandinsky-community/kandinsky-2-1-prior}}} \\
    Kandinsky 2.2 & & \footnotesize{\texttt{\href{https://huggingface.co/kandinsky-community/kandinsky-2-2-prior}{kandinsky-community/kandinsky-2-2-prior}}} \\
    Kandinsky 3 & & \footnotesize{\texttt{\href{https://huggingface.co/kandinsky-community/kandinsky-3}{kandinsky-community/kandinsky-3}}} \\
    PixArt-$\alpha$ & & \footnotesize{\texttt{\href{https://huggingface.co/PixArt-alpha/PixArt-XL-2-1024-MS}{PixArt-alpha/PixArt-XL-2-1024-MS}}} \\
    \bottomrule
  \end{tabular}
}
\vspace{-.4cm}
\end{table}

\tit{Encoding}
Encoding formats for the generated images are chosen by following the distribution of image formats in LAION-400M~\cite{schuhmann2021laion}. As a consequence, we encode roughly 91\% of the generated images in JPEG format, and the remaining 9\% with other formats (BMP, GIF, TIFF, PNG).

\tit{Negative prompts}
To enhance the quality of generated content, we also employ negative prompts. These aim at decreasing the probability of generating a specific subject or propriety. Notably, not all diffusion models accommodate negative prompts as input. Consequently, we implement this technique across all generators except for Stable Diffusion 1.4. 
A detailed list of the negative prompts employed in the generation process is reported in Table~\ref{tab:negative}. During the generation of the dataset, we apply with a probability of 0.5 five randomly sampled negative prompts.

\tit{Prompt engineering}
Moreover, prompt engineering techniques have been applied to improve the generation quality. In particular, we compose our prompts by applying the following template: 

\smallskip
\noindent``\texttt{A  [<std>] <photo type> of <prompt>, [<shot type>],[<lighting> light], [<context>], [<lens> lens]}'', 

\smallskip
\noindent where \texttt{<std>}, \texttt{<photo type>}, \texttt{<shot type>}, \texttt{<lightning>}, \texttt{<context>} and \texttt{<lens>} indicate modifiers which can be added to the prompt. Modifiers 
\begin{wrapfigure}{r}{0.52\textwidth}
\vspace{-0.5cm}
\centering
\includegraphics[width=0.98\linewidth]{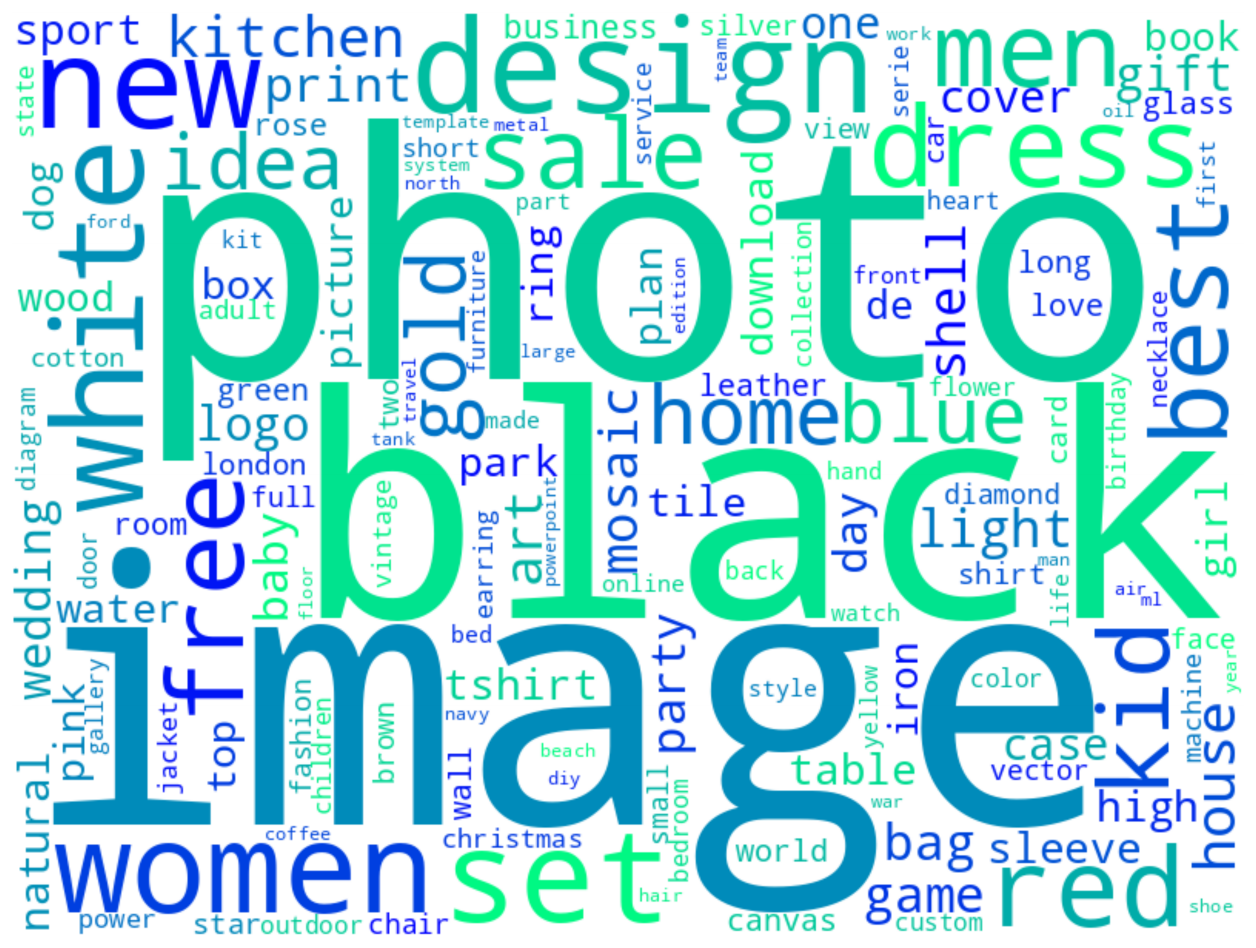}
\vspace{-0.15cm}
\caption{Word cloud obtained from a random subset of LAION-400M.}
\label{fig:wordcloud_1}
\vspace{-0.7cm}
\end{wrapfigure}
between square brackets are optional, and their presence in the prompt is sampled with a probability of 0.8. The specific modifiers are then uniformly sampled from a list of possible modifiers for each category. For a full reference, in Table~\ref{tab:modifiers} we report the complete list of modifiers for each category. 

\begin{figure*}[t]
\centering
\includegraphics[width=\linewidth]{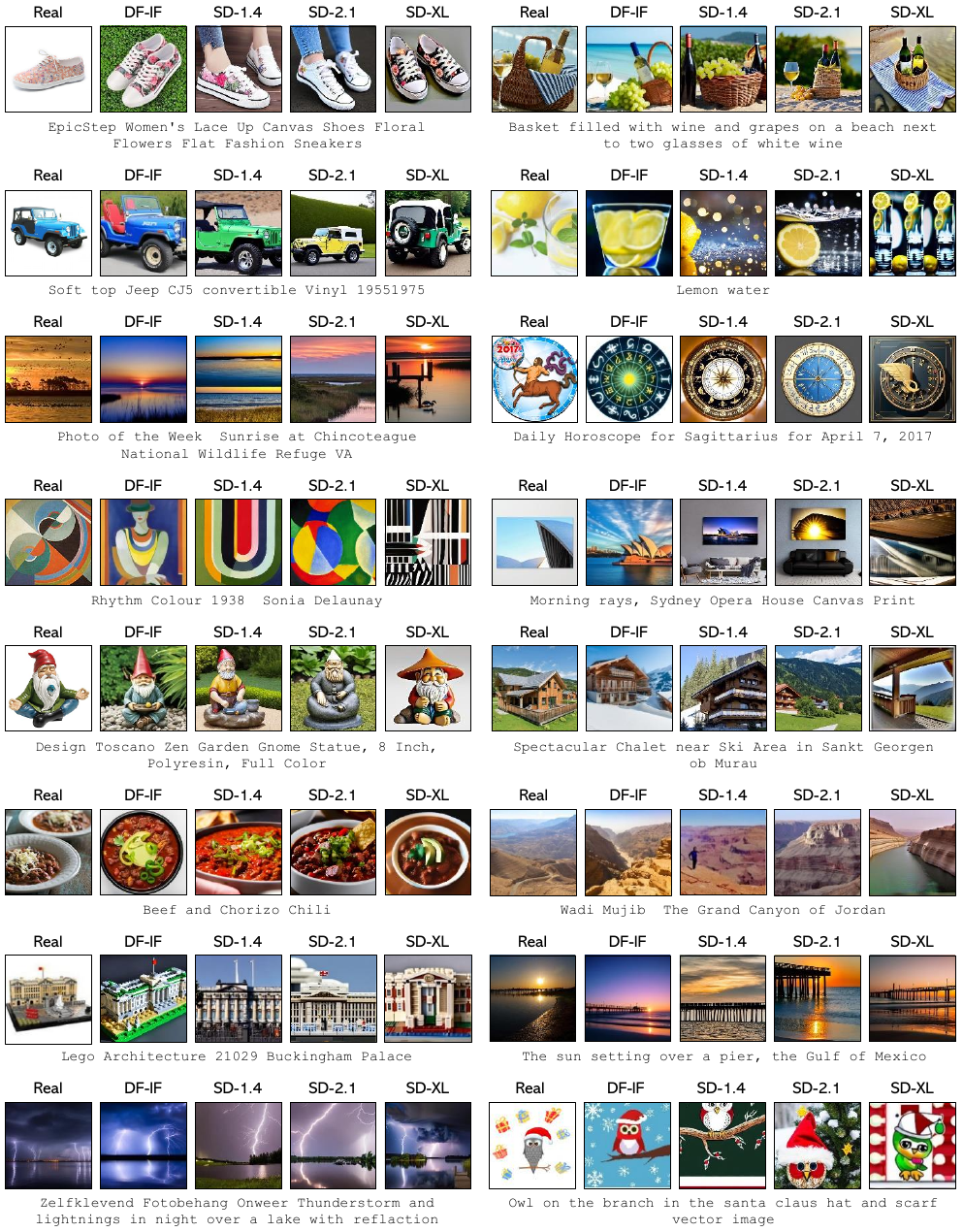}
\vspace{-0.5cm}
\caption{Qualitative samples from the \dataset dataset.}
\label{fig:dataset_supp}
\end{figure*}

\tit{Qualitative samples}
We show the word cloud distribution of LAION-400M prompts in Fig.~\ref{fig:wordcloud_1}. Finally, additional qualitative samples of the \dataset dataset are reported in Fig.~\ref{fig:dataset_supp}. For each record, we also report the textual prompt from the LAION-400M dataset without including the prompt engineering techniques previously described.

\section{Limitations and Societal Impact\vspace{-0.1cm}}
\tinytit{Limitations}
\ours is an embedding space specifically tailored for deepfake detection. However, it must be acknowledged that the model does not guarantee an infallible classification of each input image as either real or counterfeit. Indeed, as all other deepfake detection approaches, the model can potentially misclassify images, resulting in both false positives and false negatives. Furthermore, although \ours has been evaluated across a spectrum of image generators, its adaptability to new, unforeseen generators cannot be guaranteed. Future deepfakes may necessitate a fine-tuning of \ours to maintain its efficacy. 

\tit{Societal impact}
Deepfakes pose a direct threat to the integrity of information, undermining trust in digital media. Indeed, the ability to create indistinguishable fake content can reduce public confidence in visual evidence. Also, the use of deepfakes to generate imagery without the consent of the individuals involved constitutes a grave invasion of privacy and personal autonomy. Such manipulations not only violate individual rights but also have the potential to catalyze social discord and manipulate narratives related to public safety, either by propagating unfounded alarms or by detracting from legitimate warnings. Preserving the authenticity and integrity of digital content underscores the necessity of implementing rapid inference systems, a point further highlighted in Table~\ref{tab:inference_time}.

\begin{table}[t]
  \caption{List of the negative prompts employed during the generation of the \dataset dataset.}
  \label{tab:negative}
  \vspace{-0.2cm}
  \centering
  \setlength{\tabcolsep}{0.5em}
  \resizebox{0.78\linewidth}{!}{
  \begin{tabular}{llll}
    \toprule
    \rowcolor{lightgray}
    bad anatomy & blurry & fuzzy & disfigured  \\
    misshaped & mutant & deformed & bad art  \\
    \rowcolor{lightgray}
    out of frame & cgi & octane & render  \\
    3d & doll & unreal engine & unrealistic  \\
    \rowcolor{lightgray}
    retro & low quality & out of focus & inaccurate  \\
    cartoon & cartoonish & colorless & computer graphic  \\
    \rowcolor{lightgray}
    graphic & art & digital art & signature  \\
    watermark & abstract & low-resolution & unreal  \\
    \rowcolor{lightgray}
    digital & asymmetric & weird colors & canvas frame  \\
    photoshop & video game & mutated & mutation  \\
    \rowcolor{lightgray}
    poorly drawn face & bad proportions & grainy & signature  \\
    cut off & oversaturated &  &   \\
    \bottomrule
  \end{tabular}
}
  \vspace{-0.1cm}
\end{table}

\begin{table}[t]
  \centering
 \caption{List of prompt modifiers employed during the generation of the \dataset dataset.}
  \label{tab:modifiers}
  \vspace{-0.2cm}
  \setlength{\tabcolsep}{0.5em}
  \resizebox{0.85\linewidth}{!}{
  \begin{tabular}{llllll}
    \toprule
    \textbf{Shot Type} & \textbf{Lighting} & \textbf{Context} & \textbf{Lens} & \textbf{Std} & \textbf{Photo Type} \\
    \midrule
    \rowcolor{lightgray}
    close-up & soft & indoor & wide-angle & portrait & photo \\
    POV & ambient & outdoor & telephoto & ultra  detailed & image \\
    \rowcolor{lightgray}
    medium shot & ring & at night & 24mm & hyper realistic & picture  \\
    long shot & sun & studio & EF 70mm & high quality &  \\
    \rowcolor{lightgray}
     & cinematic & & bokeh & hyper detailed & \\
     & volumetric & & & photorealistic & \\
    \rowcolor{lightgray}
     & uplight & & & realistic &  \\ 
     &  &  &  & 8k &  \\
    \rowcolor{lightgray}
     &  &  &  & dramatic & \\
     &  &  &  & 4k detail & \\ 
    \bottomrule
  \end{tabular}
}
  \vspace{-0.4cm}
\end{table}

\end{document}